%% file: main.tex
\documentclass[10pt,twocolumn,letterpaper]{article}

\usepackage{cvpr}              

\input{preamble}

\definecolor{cvprblue}{rgb}{0.21,0.49,0.74}
\usepackage[pagebackref,breaklinks,colorlinks,citecolor=cvprblue]{hyperref}

\def\paperID{15259} 
\def\confName{CVPR}
\def\confYear{2024}

\title{Learning to Prompt with Text Only Supervision for Vision-Language Models}

 \author{
  Muhammad Uzair Khattak$^{1}$ \quad 
  Muhammad Ferjad Naeem$^{2}$ \quad 
  Muzammal Naseer$^{1}$ \\
  Luc Van Gool$^{2}$ \quad
  Federico Tombari$^{3,4}$
  \vspace{0.2em} \\
  $^{1}$Mohamed bin Zayed University of AI \quad 
  $^{2}$ETH Zurich \quad 
  $^{3}$TU Munich \quad 
  $^{4}$Google
}

\begin{document}
\maketitle
\input{sec/0_abstract}

\input{sec/1_intro}
\input{sec/3_finalcopy}

{
    \small
    \bibliographystyle{ieeenat_fullname}
    \bibliography{main}
}

\input{sec/X_suppl}

\end{document}

%% file: preamble.tex
\usepackage[dvipsnames]{xcolor}

\usepackage{bbding}
\usepackage{pifont}
\usepackage{wasysym}
\usepackage{amssymb}
\input{math_commands}

\usepackage{sidecap}
\newcommand{\tablestyle}[2]{\setlength{\tabcolsep}{#1}\renewcommand{\arraystretch}{#2}\centering\footnotesize}
\usepackage{float}
\definecolor{purple}{RGB}{230, 227, 254}
\definecolor{lightgreen}{RGB}{238, 252, 241}
\definecolor{lightred}{RGB}{231, 187, 187}
\definecolor{darkred}{RGB}{198, 129, 129}

\definecolor{tabhighlight}{HTML}{e5e5e5}

\newcommand{\rotbox}[1]{\rotatebox{55}{#1}}
\newcommand{\rotboxsub}[1]{\rotatebox{90}{#1}}
\usepackage{graphicx, amsmath, amssymb, caption, subcaption, multirow, overpic}
\definecolor{tabhighlight}{HTML}{e5e5e5}
\definecolor{citecolor}{HTML}{0071bc}

\definecolor{tabhighlight}{HTML}{e5e5e5}

\usepackage[dvipsnames]{xcolor}
\usepackage{xcolor,colortbl}
\usepackage{color, colortbl}
\usepackage{xcolor}
\newcommand{\tableCellHeight}{1}
\newcommand{\tabstyle}[1]{
  \setlength{\tabcolsep}{#1}
  \renewcommand{\arraystretch}{\tableCellHeight}
  \centering
  \small
}

\definecolor{grey}{RGB}{128,138,135}
\definecolor{darkgrey}{RGB}{96,96,96}

%% file: math_commands.tex
\usepackage{amsmath,amsfonts,bm}

\def\eqref#1{equation~\ref{#1}}

\def\1{\bm{1}}

\DeclareMathAlphabet{\mathsfit}{\encodingdefault}{\sfdefault}{m}{sl}
\SetMathAlphabet{\mathsfit}{bold}{\encodingdefault}{\sfdefault}{bx}{n}



%% file: sec/0_abstract.tex
\begin{abstract}
Foundational vision-language models such as CLIP are becoming a new paradigm in vision, due to their excellent generalization abilities. However, adapting these models for downstream tasks while maintaining their generalization remains a challenge. In literature, one branch of methods adapts CLIP by learning prompts using visual information. While effective, most of these works require labeled data which is not practical, and often struggle to generalize towards new datasets due to over-fitting on the source data. An alternative approach resorts to training-free methods by generating class descriptions from large language models (LLMs) and perform prompt ensembling. However, these methods often generate class specific prompts that cannot be transferred to other classes, which incur higher costs by generating LLM descriptions for each class separately. In this work, we propose to combine the strengths of these both streams of methods by learning prompts using only text data derived from LLMs. As supervised training of prompts is not trivial due to absence of images, we develop a training approach that allows prompts to extract rich contextual knowledge from LLM data. Moreover, with LLM contextual data mapped within the learned prompts, it enables zero-shot transfer of prompts to new classes and datasets potentially cutting the LLM prompt engineering cost. To the best of our knowledge, this is the first work that learns generalized prompts using text only data. We perform extensive evaluations on 4 benchmarks where our method improves over prior ensembling works while being competitive to those utilizing labeled images. Our code and pre-trained models are available at \href{https://github.com/muzairkhattak/ProText}{https://github.com/muzairkhattak/ProText}.
\vspace{-1em}
\end{abstract}

%% file: sec/1_intro.tex
\section{Introduction}
\label{sec:intro}

\input{tables/main_experiments/taxomy_comparison_table}

The Vision field is experiencing a new paradigm in its model-building approach with the emergence of foundational models \cite{radford2021learning, jia2021scaling,yu2022coca,lai2023lisa}, which are large DNNs pre-trained on web-scale data. Among these, Vision-Language models (VLMs) such as CLIP \cite{radford2021learning} stand out as the latest highlights which leverage contrastive pre-training on massive image-text pairs from the internet. During pre-training, CLIP learns to align image-text samples in a shared feature space. This allows CLIP to encode open-vocabulary concepts and generalize well to zero-shot recognition tasks.

CLIP consists of two encoders to encode image and text inputs respectively. At inference, a hand-crafted prompt such as \texttt{`a photo of a} {\textsc{cls}}' is used as the text input. Text features of classes are compared with visual feature and class with highest similarity is assigned as predicted label. Improving the quality of text templates such as adding attributes \cite{an2023more}, or class-specific details \cite{pratt2023does, jin2021good} has shown to improve CLIP performance. However, designing high-quality prompts that can best describe test image remains a key challenge, as image content is not known in advance.

In literature, numerous techniques have been proposed to adapt CLIP for downstream recognition tasks. One branch of methods \cite{zhou2022conditional,zhou2022learning, chen2022plot, huang2022unsupervised, shu2022test, lu2022prompt} treat text prompts as learnable vectors and optimize them using task-specific objectives such as cross-entropy. As prompts are learned in the embedding space, this allows them to be used with classes and datasets beyond those on which they were trained on. While effective over the baseline CLIP, most of these methods require annotated image labels to optimize the prompts which is often impractical, especially in real-world scenarios such as medical imaging, remote sensing, security, surveillance, etc. Moreover, these methods tend to overfit on few-shot source samples and struggle to retain CLIP's generalization, especially in cross-dataset settings.

Alternatively, several methods \cite{pratt2023does, menon2022visual} have adopted the training-free approach of prompt ensembling by leveraging the capabilities of Large Language Models (LLMs). Instead of using hand-crafted templates, these methods mine dataset or class specific descriptors and captions from LLMs to enrich text features. These enriched features aim to better represent content that could possibly occur in test images, leading to improvements over baseline CLIP. Although these methods do not require image information, the knowledge acquired from LLMs is mostly specific to each class and not directly transferable across unseen classes and datasets since no optimization is performed. Additionally, generating LLM descriptions for each concept separately incurs additional LLM serving and prompt engineering costs.
\begin{figure}[!t]
    \centering
    \includegraphics[width=1\columnwidth]{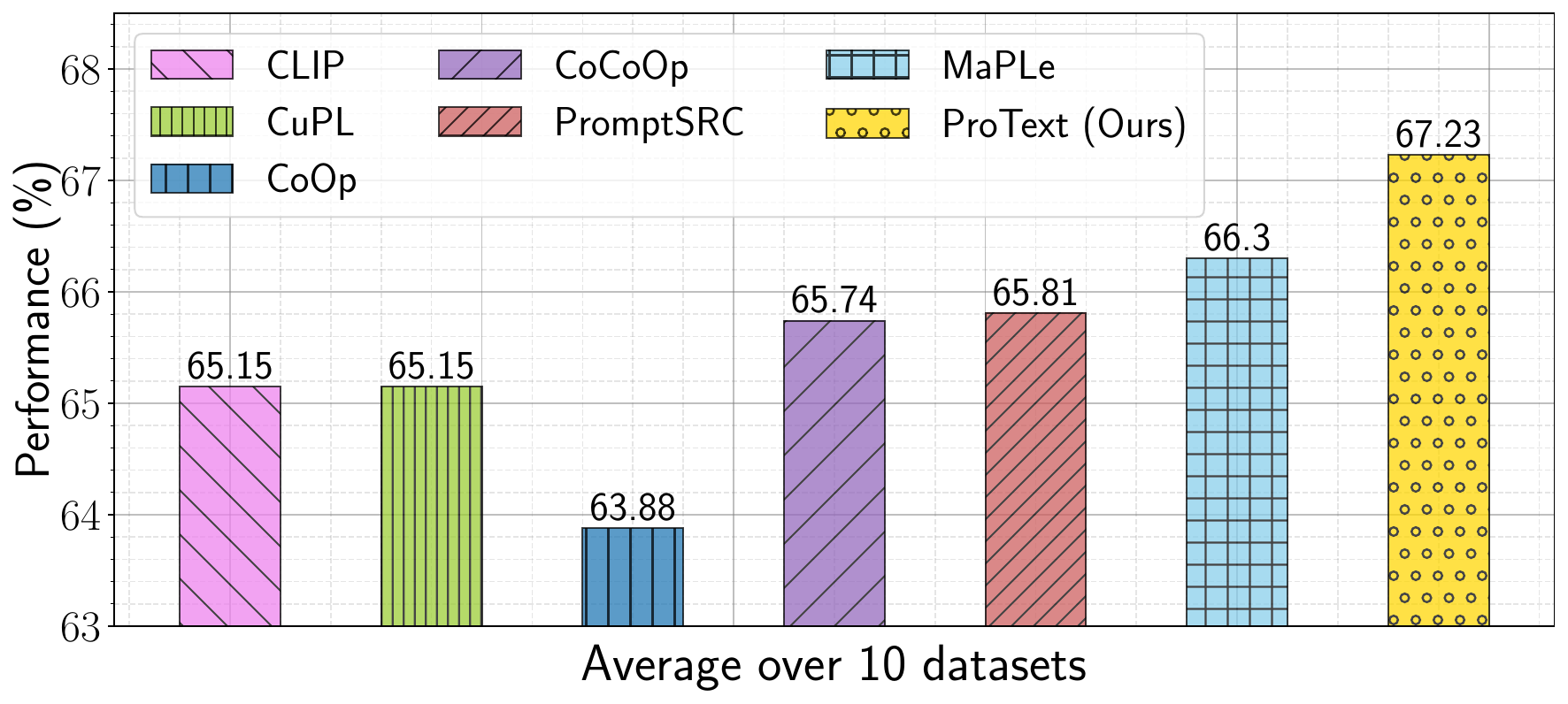}
    \vspace{-0.26in}
\caption{Without using any images for supervision, ProText with text-only training improves over CLIP, CuPL, and prior 16-shot image-supervised methods in challenging cross-dataset transfer settings. Prompt ensembling based CuPL performs same as CLIP as it cannot transfer class specific LLM templates to cross-datasets.}
  \label{fig:result_comparision}
      \vspace{-0.20in}
\end{figure}

In this work, we present a new paradigm to improve CLIP's generalization. Our motivation comes from combining the strengths of prompt learning and prompt ensembling approaches while effectively addressing their limitations. To this end, we introduce ProText: \textbf{Pro}mpt Learning with \textbf{Text}-Only Supervision. In contrast to previous methods, our approach instead proposes to learn prompts using text only data obtained from LLMs. As supervised training of prompts is not trivial due to image-free setting, we develop a novel training framework that allows prompts to learn and extract rich contextual knowledge from LLM data. Moreover, as LLM contextual knowledge is mapped within the learned prompts, it enables zero-shot transfer of prompts to new classes and datasets, potentially leading to a substantial reduction in LLM serving and prompt engineering cost. 

As shown in Tab. \ref{tab:first_table}, our approach is different from prior methods as it does not require image samples to learn prompts, in addition the adapted CLIP transfers well to unseen classes and datasets, therefore addressing a key limitation of LLM-based prompt ensembling techniques. We demonstrate the effectiveness of ProText by performing extensive evaluations on 4 benchmarks. On challenging cross-dataset transfer setting, ProText without using any visual information achieves an average gain of +2.08\% over CLIP while surpassing the performance of previous best image-supervised prompt learning method MaPLe \cite{khattak2023maple} by +0.93\% (Fig. \ref{fig:result_comparision}). Further, ProText with text-only supervision performs competitively against prior methods in domain generalization, base-to-novel class, and text-only supervised setting. Our main contributions are summarized as follows:
\begin{itemize}\setlength{\itemsep}{0em}
\item We present a new approach for prompt learning in CLIP using text-only supervision. Our method harmonically combines the strengths of prompt learning and prompt ensembling methods to improve CLIP's generalization.
\item To optimize prompts with text-only data, we develop a training approach that allows prompts to learn a mapping by extracting rich contextual information from LLM data. 
\item As LLM contextual knowledge is mapped within the learned prompts, this enables prompts to be directly used with new classes and datasets potentially cutting the additional LLM serving and prompt engineering cost.
\item We validate the effectiveness of our method through extensive experiments across four benchmarks. Our TextPro approach improves the generalization of CLIP across various settings and fares competitive to approaches that explicitly use labeled image samples for training.
\end{itemize}

\section{Related Work }
\textbf{Foundational Vision-Language models (VLMs).}
VLMs \cite{radford2021learning, jia2021scaling, yu2022coca, yuan2021florence, yao2021filip, naeem2023silc} leverage joint image-text pretraining using internet-scale data in a self-supervised fashion. Representative VLMs like CLIP \cite{radford2021learning} and ALIGN \cite{jia2021scaling} have utilized around 400M and 1B image-text pairs during their pre-training. Using the contrastive learning objective, VLMs learn rich multi-modal features by attracting together the features of paired images and texts while repelling un-paired image-text features in a joint feature space. The resulting model learns open-vocabulary concepts interpretable through natural language suitable for various downstream discriminative vision tasks such as open-vocabulary image classification \cite{khattak2023maple, lu2022prompt, chen2022plot, zhou2022learning, naeem2022i2dformer, naeem2023i2mvformer}, detection \cite{du2022learning, zhou2022detecting, minderer2022simple, liu2023grounding, bangalath2022bridging}, and segmentation \cite{ghiasi2022scaling, li2022languagedriven, liang2023open}. Although promising, adapting VLMs effectively while maintaining their original generalization remains a crucial challenge. In this work, we propose a novel method to adapt CLIP with prompt learning through \textit{text modality} supervision to improve its performance on \textit{vision modality} tasks.

\noindent \textbf{Prompt Learning for VLMs.}
Prompt Learning \cite{chen2022plot, zhou2022conditional, zhou2022learning, lu2022prompt, derakhshani2023bayesian, shu2022test, samadh2023align} has emerged as an effective fine-tuning strategy to adapt large-scale models. This approach adds a small number of learnable embeddings along with model inputs which are optimized during training while the rest of the model is kept frozen. As the pre-trained model is unchanged during prompt learning, it has become particularly effective for VLMs such as CLIP, where maintaining the model's original generalizability is crucial. CoOp \cite{zhou2022learning} is the pioneering prompt learning method for CLIP which learns text prompt embeddings to fine-tune CLIP. CoCoOp \cite{zhou2022conditional} improves CoOp's generalization by conditioning text prompts on visual features. MaPLe \cite{khattak2023maple} proposes a multi-modal prompting framework to adapt both vision and language branches of CLIP. UPL \cite{huang2022unsupervised} adopts an unsupervised prompt learning approach to finetune CLIP. PromptSRC \cite{khattak2023self} improves prompt learning from a regularization perspective by making use of additional loss functions during training. While these methods improve baseline CLIP performance, most of them require image samples with labels, which is less practical, and generating pseudo-labels is often less effective. In contrast, we present a novel prompt learning approach that improves CLIP generalization without relying on any visual samples during training.

\noindent \textbf{Training-Free Text Prompt Enhancement.}
With the emergence of LLMs such as GPT-3 \cite{brown2020language}, several approaches \cite{menon2022visual, roth2023waffling, pratt2023does} have demonstrated their potential for improving zero-shot generalization of CLIP. Instead of using hand-crafted templates for generating class features, these methods leverage LLMs to generate high-level concepts, class descriptions, and/or attributes which are used in one form or another to produce enriched text features. DCLIP \cite{menon2022visual} generates fine-grained per-class language descriptors and ensemble its similarity with image to produce classification scores. WaffleCLIP \cite{roth2023waffling} matches DCLIP performance with random descriptors and show further gains by data-specific concepts generated via LLMs. CuPL \cite{pratt2023does} query LLMs to generate class-specific prompt descriptions for text prompt ensembling. Although effective, most of these approaches generate class-specific text data from LLMs which are not directly transferable to unseen classes and new datasets since no training is performed. On the other hand, we aim to leverage the same LLM data via novel text-only prompt learning technique which seamlessly allows the transfer of learned prompts toward unseen classes and new datasets.

\section{Method}

\begin{figure*}[!ht]
\centering
{\includegraphics[width=0.92\textwidth]{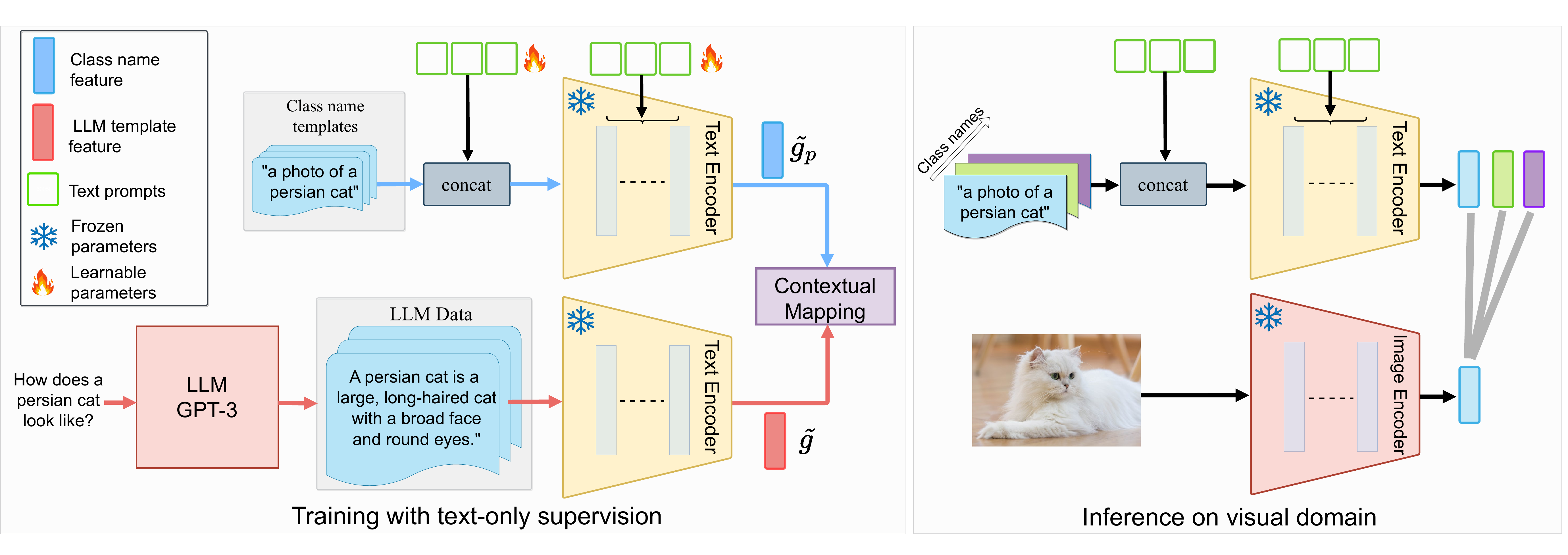}}\vspace{-0.5em}
\caption{Overview of ProText framework. \textbf{(Left)} First, diverse captions are generated for training classes using LLM like GPT-3. During training, CLIP text encoders generate \textcolor{RoyalBlue}{\textbf{prompted class-name feature}} ($\bm{\Tilde{g}_p}$) from class-name templates with learnable prompts and \textcolor{Red}{\textbf{frozen LLM template feature}} ($\bm{\Tilde{g}}$) from LLM generated templates. Next, we employ contextual mapping loss to guide learnable prompts to learn a mapping from the prompted class-name feature to the LLM template feature containing more information about the class. This allows the learned prompts to exploit internal knowledge of text encoder complemented by LLM descriptions. \textbf{(Right)} At inference, learned prompts are used with class-name templates, and the standard zero-shot CLIP inference protocol is followed. Moreover, rich contextual information from LLM descriptions mapped within the learned prompts enables its transferability to new classes and datasets.}
\label{fig:main_figure}
\vspace{-0.5em}
\end{figure*}

Given the language interpretable nature of foundational VLMs such as CLIP \cite{radford2021learning}, they are naturally suited for zero-shot recognition tasks. However, to achieve full potential of CLIP's generalization for downstream tasks, adaptation still appears to be necessary. 
Numerous approaches have since been proposed to adapt general knowledge of CLIP for user-specific downstream tasks. One line of methods adopts prompt learning \cite{lu2022prompt, khattak2023maple, zhou2022conditional,zhou2022learning} to re-purpose CLIP features for downstream data. While effective, most of them require image samples with labels to learn the prompts, which is a hard requirement to meet. Another line of methods adopts training-free prompt ensembling techniques \cite{pratt2023does, menon2022visual, roth2023waffling} with the help of LLMs. Although ensembling-based approaches do not require image information, the majority of these works generate class-specific LLM prompts that are not directly transferable to new classes and datasets.

To this end, we present a new paradigm for learning generalized transferable prompts for VLMs using text-only supervision. Our proposed adaptation framework, ProText: \textbf{Pro}mpt Learning with \textbf{Text} only supervision aims to address the challenges of existing approaches by learning \textit{transferable} prompts without relying on images. Fig. \ref{fig:main_figure} shows our ProText framework. First, we curate text-only LLM template data using class names of a given dataset and a LLM such as GPT-3 \cite{brown2020language}. As a text-supervised approach, ProText only requires CLIP text encoders during training. Specifically, we employ one frozen encoder with learnable prompts and a second frozen encoder without learnable prompts. Learnable prompts with class-name templates are input to the prompted text encoder to obtain the class-name template feature, and a frozen text encoder generates LLM template feature from its description obtained from LLM data. Next, we employ a contextual mapping training objective which maps class-name template feature to the LLM template feature. Contextual mapping allows the prompts to learn a mapping function that embeds rich contextual knowledge from LLM data within the prompt vectors. As prompts are learned in the embedding space, they are directly compatible with new classes and datasets. At inference, the learned prompts are shipped with CLIP model for standard zero-shot CLIP inference for visual recognition. 

Below we explain our proposed approach in detail. We first revisit CLIP and previous methods including Prompt Learning and Prompt Ensembling via LLMs in \autoref{sec:prompt_learning_recap} and then we present our ProText approach in \autoref{sec:protext_approach}.

\subsection{Preliminaries}
\label{sec:prompt_learning_recap}
\textbf{Contrastive Language-Image Pre-training (CLIP).} CLIP consist of an image encoder ${f}$ and a text encoder ${g}$ which maps image and text input into visual and textual feature respectively. We denote CLIP parameters as ${\theta}_{\mathtt{CLIP}} = \{\theta_{f}, \theta_{g} \}$ where $\theta_{f}$ and $\theta_{g}$ refer to the image and text encoder parameters, respectively. Input image $\bm{X}$ is divided into $M$ patches which are linearly projected to produce patch tokens and a learnable class token ${\textsc{cls}}$ is prepended resulting in the final sequence as $\bm{\Tilde{X}}= \{\textsc{cls}, \bm{e}_{1}, \bm{e}_{2}, \cdots, \bm{e}_{M}\}$. The image encoder ${f}$ encodes the input patches via multiple transformer blocks to produce a latent visual feature representation $\bm{\Tilde{f}} = f(\bm{\Tilde{X}}, \theta_{f})$, where $\bm{\Tilde{f}} \in  \mathbb{R}^{d}$. Next, the corresponding class label ${y}$ is embedded in a text template, such as \texttt{`a photo of a} [{\textsc{class}}]' which can be formulated as $\bm{\Tilde{Y}}=\{\textsc{SOS}, \bm{t}_{1}, \bm{t}_{2}, \cdots, \bm{t}_{L}, \bm{c}_{k}, \textsc{EOS}\}$. Here $\{\bm{t}_l|_{l=1}^{L}\}$ and $\bm{c}_{k}$ are the word embeddings corresponding to the text template
and the label $y$, respectively while $\textsc{SOS}$  and  $\textsc{EOS}$ are the learnable start and end token embeddings. The text encoder ${g}$ encodes $\bm{\Tilde{Y}}$ via multiple transformer blocks to produce the latent text feature as $\bm{\Tilde{g}} = g(\bm{\Tilde{Y}}, \theta_{g})$, where $\bm{\Tilde{g}} \in  \mathbb{R}^{d}$. For zero-shot inference, text features of text template 
with class labels $\{1, 2, \cdots, C\}$ are matched with image feature $\bm{\Tilde{f}}$ as $\frac{\mathtt{exp}(\mathtt{sim}(\bm{\Tilde{g}} \cdot \bm{\Tilde{f}})\tau)}{\sum_{i=1}^{C}\mathtt{exp}(\mathtt{sim}(\bm{\Tilde{g_i}} \cdot \bm{\Tilde{f}})\tau)}$, where $\mathtt{sim}()$ denotes the cosine similarity and $\tau$ is the temperature.

\noindent  \textbf{Prompt Learning with CLIP.}
Being a parameter efficient tuning method, prompt learning has emerged as a popular technique to adapt vision-language models like CLIP. Since most of the model is kept frozen during adaptation, prompt learning aims to reduce overfitting. Learnable prompts are appended either at the image side \cite{bahng2022visual}, text encoder side \cite{zhou2022conditional, zhou2022learning}, or both sides. In this work, we learn hierarchical prompts at the text encoder named Deep Language Prompting (DLP) \cite{khattak2023maple} formulated as follows.

$T$ learnable language prompts  $\bm{P_{t}} = \{\bm{p_t}^1,\bm{p_t}^2, \cdots, \bm{p_t}^T\}$ are appended with text input tokens, resulting in $\bm{\Tilde{Y}_p}=\{\textsc{SOS}, \bm{P_{t}}, \bm{t}_{1}, \bm{t}_{2}, \cdots, \bm{t}_{L}, \bm{c}_{k}, \textsc{EOS}\}$. The text encoder processes $\bm{\Tilde{Y}_p}$ and prompted text feature is obtained as $\bm{\Tilde{g}_p} = g(\bm{\Tilde{Y}_p}, \theta_{g})$. We use deep prompting which learns hierarchical prompts at subsequent transformer blocks of text encoder. Visual feature $\bm{\Tilde{f}}$ is obtained without utilizing learnable prompts. To adapt CLIP on image classification task on dataset $\mathcal{D}$, prompts $\bm{P_t}$ are optimized in a supervised fashion using labeled image samples with cross-entropy loss, $\mathcal{L_{\text{CE}}}$.
\begin{align}
\label{eq:LCE}
    \mathcal{L_{\text{CE}}} = \text{arg}&\min_{\bm{P_t}}\mathbb{E}_{(\bm{X}, {y})\sim\mathcal{D}} \, \mathcal{L} (\text{sim}(\bm{\Tilde{f}},\bm{\Tilde{g}_p}), y).
\end{align}

\noindent \textbf{Prompt Ensembling with LLM descriptions.}
Several methods have recently proposed to adapt CLIP via training-free prompt ensembling techniques. The majority of these approaches leverage the capabilities of LLMs to mine rich descriptions, attributes, or high-level concepts of class names. The corresponding text features are either averaged \cite{pratt2023does} or the similarity score of each attribute with the image is calculated to obtain classification scores \cite{roth2023waffling} \cite{menon2022visual}. 

In this work, we focus our comparison with a strong ensembling baseline CuPL \cite{pratt2023does}. Specifically, a Large Language Model $\mathcal{F}$ such as GPT-3 \cite{brown2020language} is used to generate 
class-specific descriptions for class labels $\{1, 2, \cdots, C\}$ using queries such as \textit{`How does a \textsc{class} look like'}. Text features of the same class description are averaged together, which serves as the ensembled text features. Finally, zero-shot inference is performed with those ensembled text features.

\subsection{Prompt Learning with Text-Only Supervision}
\label{sec:protext_approach}

While image-supervised prompt learning and LLM-based prompt ensembling methods have proven effective in adapting CLIP, they face notable challenges as outlined below.

\noindent \textbf{Visual data dependency.}
Existing prompt learning methods require visual samples with labels to optimize prompts using Eq. \ref{eq:LCE}. However, collecting samples and labels is difficult in critical scenarios like medical images, remote sensing, and surveillance. Pseudo-labels alleviate label dependency but they are often less effective. Furthermore, these methods tend to overfit CLIP to source data distributions and compromise generalization across cross-datasets. For instance, CoOp utilizing labeled source samples reduces average CLIP performance by 1.27\% on 10 cross-datasets.

\noindent \textbf{LLM Prompts transferabilty limitation.}
LLM-based prompt ensembling approaches like CuPL \cite{pratt2023does} generate class-specific LLM descriptions that cannot be directly transferred to unseen classes and datasets. While open-source LLMs exhibit lower performance, proprietary ones such as GPT-3 are required for generating data for new classes and datasets leading to additional serving costs.

Our work aims to address the aforementioned limitations within a unified framework. Below we detail our strategy for curating text-to-text data via LLMs for training, followed by our text-only prompt learning framework.

\subsubsection{Text-Only LLM data for Prompt Learning}
\label{text_data_curation_para}
As discussed in \autoref{sec:prompt_learning_recap}, optimizing prompts for downstream datasets typically requires image-labels pairs. Since we explicitly aim to bypass this requirement, we first leverage LLMs to curate text data for prompt learning which consists of text inputs and text outputs. 
Given a set of classes $\{c_i\}_{i=1}^C$, we prepare text inputs $\{L_{\texttt{inputs}}^i\}_{i=1}^C$ by wrapping each class name in a standard hand-written text template,
\begin{align*}
L_{\texttt{inputs}}^i = \texttt{`a photo of a } c_i \texttt{'} .
\end{align*}

Next, we prepare text outputs corresponding to the $L_{\texttt{inputs}}$. Specifically, we query GPT-3 model to generate detailed descriptions for each class name $c_i$. Similar to CuPL \cite{pratt2023does}, we prompt GPT-3 with different queries $Q$ conditioned on class names such as \textit{`How does a $c_i$ look like?'} and \textit{`How can you identify a $c_i$?"} to obtain text outputs,
\begin{align*}
L_{\texttt{outputs}}^i = \mathcal{F}(Q|c_i).
\end{align*}

Similar to \cite{pratt2023does}, we generate $M$ text outputs per query $Q$ and use $N$ different queries, resulting in $M \times N$ text outputs per class category. We associate all $L_{\texttt{outputs}}$ with the corresponding single $L_{\texttt{inputs}}$ for each class $c_i$. As LLMs are pre-trained on internet-scale text corpora, they possess the capability of generating very diverse and high-quality descriptions and captions for different class categories which results in high-quality text outputs.
Finally we combine $L_{\texttt{inputs}}$ and $L_{\texttt{outputs}}$ to create LLM based text-to-text data for text only prompt learning, $\mathcal{D}_{ \textsc{prompt}} = \{L_{\texttt{inputs}}^{i},  L_{\texttt{outputs}}^{i}\}_{i=1}^{M\times{N}\times{C}}$.
We refer the readers to supplementary for additional details on the choice of LLM prompts and examples of $\mathcal{D}_{ \textsc{prompt}}$.
\subsubsection{Contextual mapping with Prompt Learning}
\label{contextual_mapping_para}
To leverage LLM text-to-text data $\mathcal{D}_{ \textsc{prompt}}$ for learning generalized transferable prompts, we propose a contextual mapping strategy that effectively learns a mapping function that maps standard class name templates such as $\texttt{`a photo of a } c_i \texttt{'}$ to the text feature
generated from a LLM description which contains more information about the class $c_i$. In other words, contextual mapping allows learnable prompts to map $L_{\texttt{inputs}}$ to $L_{\texttt{outputs}}$ in the text feature space of CLIP. The mapping function is realized in the form
of learnable prompt vectors, which we found to be more effective in our ablations as compared to other techniques such as adapters via linear projection and MLP.

For an $i_{\text{th}}$ training sample from $\mathcal{D}_{ \textsc{prompt}}$ consisting of a text-to-text pair $\{L_{\texttt{inputs}},  L_{\texttt{outputs}}\}_{i}$, we obtain prompted class-name feature $\bm{\Tilde{g}}_p$ for $L_{\texttt{inputs}}^i$ using learnable prompts and frozen LLM feature $\bm{\Tilde{g}}$ for $L_{\texttt{outputs}}^i$ without the prompt vectors within the pre-trained latent space of CLIP. We then impose a contextual mapping constraint between $\bm{\Tilde{g}}_p$ and $\bm{\Tilde{g}}$ text features as follows,
\begin{align}
\label{eq:contextual_mapping}
 \mathcal{L_{\texttt{mapping}}} = {\frac{1}{d}}\sum_{i=1}^{d}||\bm{\Tilde{g}_{p}} - \bm{\Tilde{g}}||_{2}^{2}.
\end{align}
As shown above, we utilize MSE loss objective to enforce contextual mapping from $L_{\texttt{inputs}}^i$ to $L_{\texttt{outputs}}^i$. We study other choices of consistency objectives in our ablations (\autoref{sec:ablation_studies}).

\noindent \textbf{Motivation for $\mathcal{L_{\texttt{mapping}}.}$} 
Contextual mapping objective allows learnable prompts to exploit internal knowledge of text encoder of CLIP to generate rich contextual features aligned with the LLM descriptions ($L_{\texttt{outputs}}^i$) for a given class. This strategy effectively learns prompts without using any visual information and when trained using all training classes together, it enables prompts to capture versatile and generalized context from the LLM descriptions. These context-aware prompts become adaptable for use with any dataset and effectively enable the transferability of class-specific LLM descriptions to unseen classes and datasets. Consequently, this substantially reduces the per-dataset overhead associated with LLM serving and prompt engineering.

\noindent \textbf{Inference.}
 Once text prompt vectors are optimized through our TextPro framework in the text domain, they become ready to be shipped with CLIP for downstream visual domain inference with a standard zero-shot CLIP inference setup. As shown in Fig. \ref{fig:main_figure} (right), the learned prompts $\bm{P_t}$ are fused with each given class name to produce prompted text features $\{\bm{\Tilde{g}_{p}}\}_{i=1}^C$. Finally, zero-shot inference is performed with the prompted text features and the input image feature $\bm{\Tilde{f}}$ to produce classification scores on test images.
\section{Experiments}
\subsection{Evaluation settings}
We perform evaluations in 4 benchmark settings. Prompt ensembling methods and ProText utilize text-only LLM data for adapting CLIP while image-supervised prompt learning methods use image-label pairs for training.

\noindent \textbf{Base-to-Novel Generalization.} This setting evaluates the generalization of methods within a dataset. Following previous methods \cite{zhou2022learning, zhou2022conditional}, we split each dataset into base and novel classes. Models are trained on base classes and evaluated on the test set of base and novel classes respectively.

\noindent \textbf{Cross-dataset transfer.} This setting evaluates the generalization ability of models trained on ImageNet-1k \cite{deng2009imagenet} source dataset by directly transferring it on cross-datasets.

\noindent \textbf{Domain Generalization.} We evaluate the robustness of different methods on out-of-distribution datasets. We train models on the ImageNet-1k source dataset and evaluate its performance on four ImageNet variants with domain shifts.

\noindent \textbf{Supervised setting.} We provide performance comparison of ProText with CuPL\cite{pratt2023does} with text-only data per dataset. 

\input{tables/main_experiments/supervised_comparison_different_text_data}
\input{tables/main_experiments/base_to_novel}
\noindent\textbf{Datasets.} For the aforementioned benchmarks, we use same datasets as followed by previous works \cite{zhou2022learning, zhou2022conditional, khattak2023maple, khattak2023self}.  For cross-dataset transfer, domain generalization, and base-to-novel generalization settings, we use 11 image datasets that cover multiple recognition tasks. These includes ImageNet~\cite{deng2009imagenet} and Caltech101~\cite{fei2004learning} which contains generic objects; OxfordPets~\cite{parkhi2012cats}, StanfordCars~\cite{krause20133d}, Flowers102~\cite{nilsback2008automated}, Food101~\cite{bossard2014food}, and FGVCAircraft~\cite{maji2013fine} for fine-grained classification, SUN397~\cite{xiao2010sun} for scene recognition, UCF101~\cite{soomro2012ucf101} for action recognition, DTD~\cite{cimpoi2014describing} for texture classification, and EuroSAT~\cite{helber2019eurosat} for satellite images categorization. For domain generalization setting, we train models on ImageNet~\cite{deng2009imagenet} as a source dataset and use ImageNet-A~\cite{hendrycks2021natural}, ImageNet-R~\cite{hendrycks2021many}, ImageNet-Sketch~\cite{wang2019learning} and ImageNetV2~\cite{recht2019imagenet} for out of distribution dataset evaluation.

\noindent\textbf{Implementation details.}
We use a publically available pretrained ViT-B/16 CLIP model from OpenAI \cite{radford2021learning}. We train ProText with Deep Language Prompting in the first 9 transformer blocks of the CLIP text encoder.
For cross-dataset transfer and domain generalization setting, we train ProText using $T=4$ and $T=16$ language prompts with 10 and 200 epochs respectively. Similar to \cite{wang2019learning}, ProText and zero-shot CLIP use additional concepts where available with its prompts such as $\texttt{`a photo of a CLS, a type of flower} \texttt{'}$ for OxfordFlowers \cite{nilsback2008automated}. For base-to-novel and supervised text-only settings, ProText uses optimal prompt length and epoch configuration for each dataset. Optimal training configuration is obtained through hyper-parameter search on validation split of datasets. To generate text-only data, we utilize GPT-3 DaVinci-002 model \cite{brown2020language} and generate class-specific descriptions using the LLM prompts provided by CuPL \cite{pratt2023does}. We use publicly available CuPL data and generate descriptions for datasets not provided by CuPL. AdamW optimizer is used with 5 warm-up epochs for training. We use a single 16-GB V100 to train our models.  Refer to supplementary material for additional implementation details.

\subsection{Effectiveness of Text-Only Supervision}
We first present an ablation to motivate our approach of learning prompts with text-only supervision. We train ProText with 3 types of text data and evaluate performance on ImageNet-1k \cite{deng2009imagenet}. ProText-Attribute uses 46 templates from \cite{an2023more} which corresponds to common image attributes such as rotation, blurriness, etc. ProText-80 is trained on standard 80 templates provided by CLIP \cite{radford2021learning} and ProText-CuPL is trained on class-specific LLM data employed by our main baseline CuPL \cite{pratt2023does} for its ensembling approach. 

In  Tab. \ref{table:different_text_data_motivation}, we compare ProText with CLIP and recent LLM-based ensembling methods. Prompt ensembling with attribute templates and 80 templates improves over CLIP single template result. Among the LLM-based ensembling methods, CuPL provide highest performance of 69.62\%. In contrast, ProText uses a learning-based approach and shows competitive performance against prompt ensembling methods using the same text data. ProText-Attribute provides  gain of 0.45\% over CLIP-Attribute while roughly maintaining its performance against CLIP-80. When equipped with CuPL LLM text-data, ProText surpasses CuPL by 0.60\% leading to highest performance against all methods. These results motivate our approach that instead of prompt ensembling, one can achieve competitive results by utilizing the same available text data to learn prompts. Next, we demonstrate the generalization of ProText such that the learned prompts transfer well across new classes and datasets.

\subsection{Base to novel class generalization}
We now present results in base-to-novel class generalization setting where training data for only base classes are available and the model is evaluated on both base and novel classes. For CuPL \cite{pratt2023does}, we use base-class LLM templates for base classes and zero-shot CLIP results for its novel classes. For ProText, we use base-class LLM templates for training and transfer the learned prompts for novel classes.

\input{tables/main_experiments/cross_datasets}
\input{tables/main_experiments/domain_generalization}
\input{tables/main_experiments/supervised_training_complete}
Results are shown in Tab. \ref{tab:base-to-new}.  CuPL outperforms zero-shot CLIP on base classes while maintaining its performance on novel classes as LLM prompts for new classes are not available. ProText shows consistent improvements over CuPL on base classes for 11 datasets. Furthermore, with the same LLM base-class data as CuPL, ProText effectively transfers learned prompts towards novel classes and improves CLIP and CuPL novel class performance by 2.76\% averaged across 11 datasets. This shows the advantage of ProText prompts to benefit unseen class performance potentially reducing the LLM prompt serving cost by half.

\subsection{Cross-dataset transfer}
\label{main:cross-dataset}
In cross-dataset transfer setting, we compare ProText with CLIP \cite{radford2021learning}, CuPL \cite{pratt2023does}, and image-supervised prompt learning methods. Since class-specific ImageNet LLM prompts limit its transfer to other datasets in CuPL, we assign CLIP results to CuPL for cross-datasets. Image-supervised methods \cite{zhou2022conditional, zhou2022learning, khattak2023maple, khattak2023self} are trained with 16-shot ImageNet data.

We show our main comparison results in Tab. \ref{tab:xd}. CuPL improves ImageNet performance of CLIP by ensembling ImageNet LLM prompts, while its cross-dataset results remain the same as CLIP. In contrast, ProText effectively addresses the transferability challenges of CuPL using generalized prompts trained with the same ImageNet LLM data. 
Since ProText allows generalization to unseen datasets, these learned prompts can directly be used with CLIP for cross-datasets leading to absolute average gains of +2.1\% against CLIP and CuPL. 
With ProText, one can notably reduce proprietary LLM serving and prompt engineering costs as prompts learned on one dataset are effectively transferable to other datasets. We next compare ProText with strong 16-shot image-supervised methods. Without using any visual samples, ProText demonstrates effective generalization on cross-datasets and consistently surpasses previous state-of-the-art MaPLe on 9/10 datasets leading to the highest average accuracy of 67.23\%. This highlights that text-only methods like ProText  can lead to better generalization of CLIP as compared to image-supervised methods which tend to overfit on the source sample distributions.
 
\subsection{Domain generalization experiments}
We present the results for domain generalization task in Table \ref{tab:robustness}. As the domain shift variants of ImageNet share class names with ImageNet, CuPL employs prompt ensembling for each dataset and provides an average gain of +2.84\% over CLIP. In contrast, ProText with learned prompts shows an additional gain of +0.44\% against CuPL averaged over 4 datasets. Moreover, ProText fairs competitively with image-supervised methods by showing consistent improvements over CoOp, CoCoOp, and MaPLe. These results suggest that text-only supervision methods like ProText can serve as an effective alternative to improve the robustness of VLMs when no visual information is available for training.

\subsection{Supervised text-only training}
In this setting, we compare ProText with CuPL for each dataset trained on LLM template data and the results are shown in Tab. \ref{tab:few_shot_experiments}. While utilizing the same LLM data, ProText achieves consistent improvements over CuPL on 10/11 datasets with an average gain of +0.59\%. This reflects the generalization of the ProText approach across various diverse image datasets where it better utilizes LLM data within the learned prompts. We also compare ProText with image-supervised methods and observe that ProText fares competitively with approaches utilizing up to 2-shot samples for training. This shows ProText as a potential alternative to image-supervised methods in extremely low-data regimes. Refer to supplementary for additional results.

\subsection{Ablative analysis}
\label{sec:ablation_studies}

\begin{figure}[!t]
    \centering
    \includegraphics[width=0.95\columnwidth]{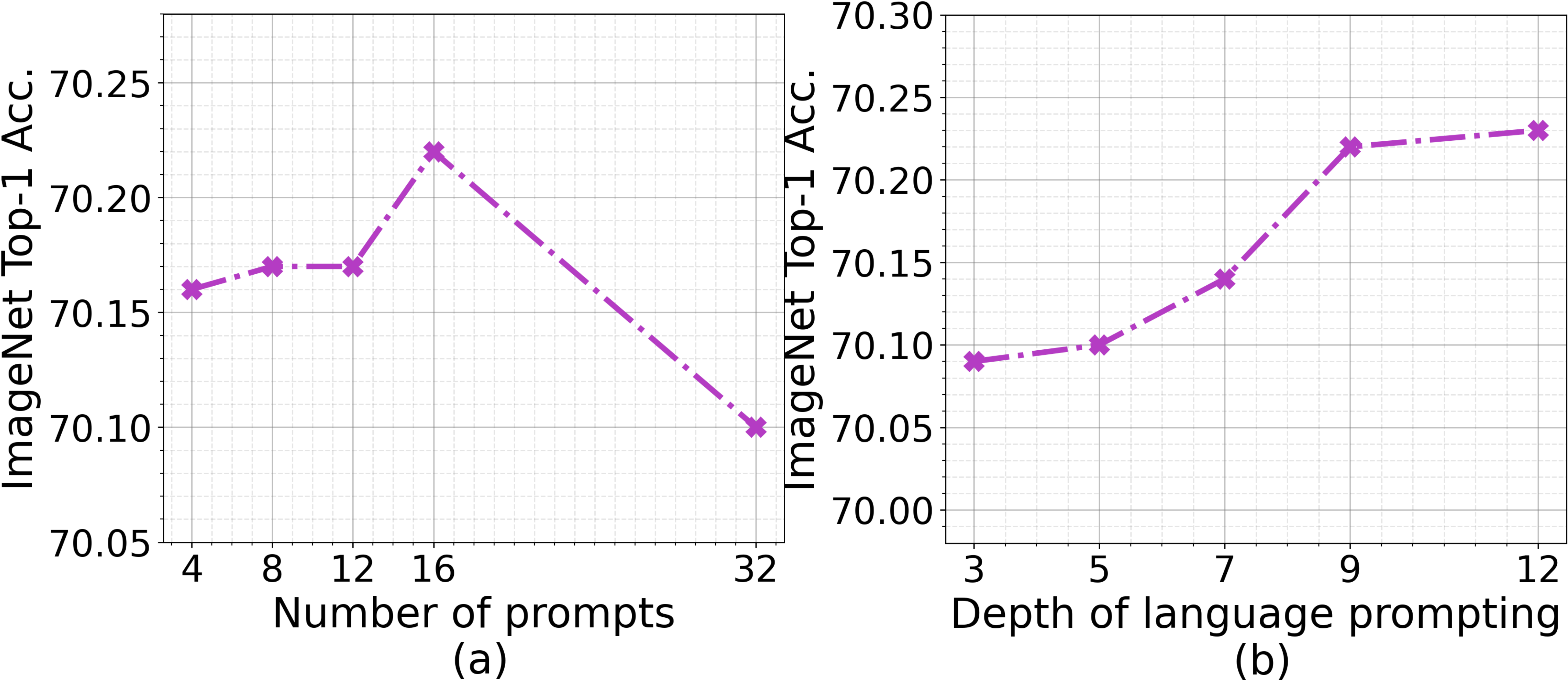}
\caption{Ablation: Prompt length (left) and prompt depth (right).}
  \label{fig:ctx_templates_ablation}
      \vspace{-0.5em}
\end{figure}

\input{tables/ablations/loss_ablation}
\input{tables/ablations/LLM_ablation}
\input{tables/ablations/adaptor_ablation}
\input{tables/main_experiments/prediction_score_analysis}
\noindent \textbf{On understanding ProText prompts.} In Table. \ref{tab:confidence_analysis}, we present average confidence scores obtained from ProText logits trained on ImageNet-1k text data when applied to cross-datasets. Compared to CLIP, ProText exhibits increased confidence scores for correct classes across various datasets, while marginally decreasing confidence scores for incorrect classes. This suggests that the prompts learned on ImageNet-1k provide complementary and transferable contextual cues, leading to improved results. We conjecture that ProText prompts potentially improve the classification of test samples situated near the decision boundary due to higher confidence for correct classes. Refer to the supplementary section for qualitative and additional analysis.

\noindent \textbf{Loss metric in contextual mapping.} We ablate on choice of loss used for the contextual mapping module in Tab. \ref{tab:matching_loss_ablations}. Distance-based losses improve over contrastive loss. We conjecture that contrastive loss treats samples of same class labels in a same batch as negatives leading to noisy training.

\noindent  \textbf{Choice of LLM for generating text data.} ProText by default uses GPT-3 \cite{brown2020language} LLM to obtain text templates for training. Here we ablate on an open-source Alpaca \cite{taori2023stanford} model as an alternative choice. As shown in Tab. \ref{tab:LLM_ablation}, ProText with Alpaca templates performs worse than ProText-80 template and ProText-GPT-3. We observed that Alpaca templates are often noisy while GPT-3 descriptions contain more enriched class details which results in better performance.

\noindent  \textbf{Prompt learning verses adapter.} While ProText employs prompt learning to learn contextual mapping from LLM templates, here ablations on adapters in Tab. \ref{tab:adaptor_ablation}. Similar to \cite{gao2023clip}, we attach adapter at the output of CLIP text encoder. Adapters perform lower as compared to prompting. We conjecture that adapter completely transforms text features and loses CLIP generalization. In contrast, prompt learning append learnable vectors with CLIP text input without significant replacement and learns effective mapping function.

\begin{figure}[!t]
    \centering
    \includegraphics[width=0.95\columnwidth]{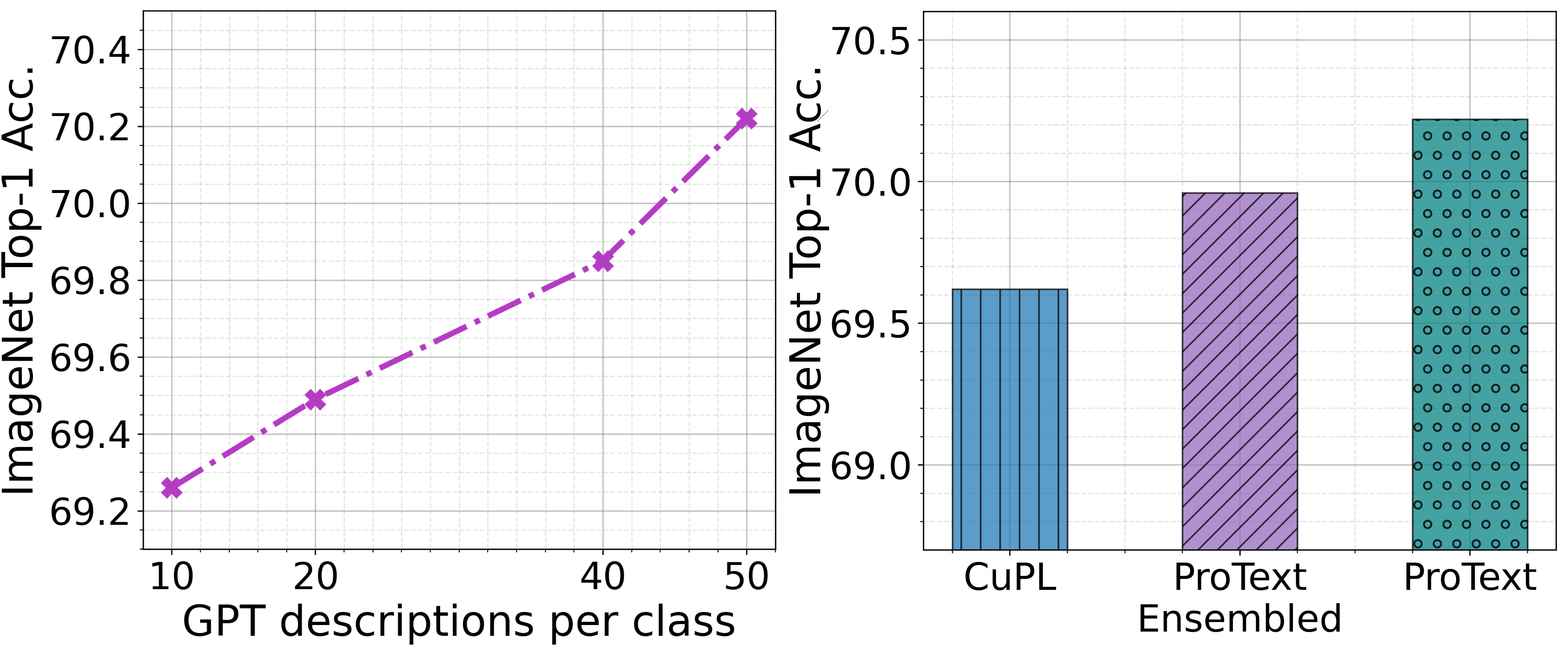}
\caption{(Left) Effect of LLM data size on performance. (Right) Ablation on ensembling LLM descriptions for training ProText.}
  \label{fig:data_size_ablation}
   \vspace{-1.6em}
\end{figure}

\noindent \textbf{Training data size for text-supervision.}
To assess the effect of LLM template data size on ProText, we ablate on the number of descriptions per class in Fig. \ref{fig:data_size_ablation} (left). Increasing descriptions for each class consistently improves the results. This suggests that we could further boost ProText performance as quality and size of text data increases. 

\noindent \textbf{Ensembling in ProText training.} ProText uses multiple descriptions per class and enforce mapping of class-name template feature to feature of each LLM description for that class. We conduct an alternative experiment by ensembling a single feature from multiple LLM descriptions per class and enforce mapping on ensembled LLM feature. As shown in Fig. \ref{fig:data_size_ablation} (right), ProText-ensembled performs lower than ProText with individual samples. 
We conjecture that learning on each description allows the model to utilize additional context present in each description. Ensembling can potentially mask out less frequent details available in text.

\noindent \textbf{Prompt length and prompt depth.}
Fig. \ref{fig:ctx_templates_ablation} (left) shows the effect of prompt length for training ProText. Setting prompt length to 16 leads to optimal performance. Fig. \ref{fig:ctx_templates_ablation} (right) shows the effect of prompt depth on final performance where prompt depth of 9 shows optimal results.

%% file: tables/main_experiments/taxomy_comparison_table.tex
\begin{table}[t!]
    \centering
    \setlength{\tabcolsep}{-0.2mm}{
    \resizebox{1\linewidth}{!}{
    \begin{tabular}{c lcc}
    \toprule
    & \multirow{2}{*}{~~~~~~Method}  & Do not require & Transfer to  \\
      & &  images & unseen datasets  \\
            \midrule

    &  ~~~~~~CoOp \cite{zhou2022learning}  & \ding{56} & \ding{51} \\
   Prompt learning & ~~~~~~CoCoOp \cite{zhou2022conditional} &  \ding{56} &  \ding{51}\\
    methods & ~~~~~~MaPLe \cite{khattak2023maple} &	 \ding{56} &  \ding{51} \\
    & ~~~~~~PromptSRC \cite{khattak2023self}& \ding{56} &  \ding{51} \\
    \midrule
        &  ~~~~~~DCLIP \cite{menon2022visual}  & \ding{51} & \ding{56} \\
   ~Prompt ensembling & ~~~~~~WaffleCLIP-Concept \cite{roth2023waffling} &  \ding{51} &  \ding{56}\\
    methods (LLM) & ~~~~~~CuPL \cite{pratt2023does} &	 \ding{51} &  \ding{56} \\
    \midrule
    \rowcolor{tabhighlight}
    & ~~~~~~ProText (Ours)  &	\ding{51} &	\ding{51}   \\
    \bottomrule
    \end{tabular}
    }}
    \caption{Existing methods improve CLIP's generalization by learning prompts with image supervision or using non-transferable prompt ensembling with LLM knowledge. In contrast, our approach, ProText, effectively learns prompts with text-only supervision which are transferable to new datasets and classes.
    }
    \label{tab:first_table}
    \vspace{-0.20in}
\end{table}

%% file: tables/main_experiments/supervised_comparison_different_text_data.tex
\begin{SCtable}[][!t]
    \centering
    \setlength{\tabcolsep}{1mm}{
    \resizebox{0.62\linewidth}{!}{
    \begin{tabular}{l c}
    \toprule
    Method  & ImageNet Acc.\\
            \midrule
    1: CLIP (ICML'21) & 66.72\\
    2: CLIP-Attribute & 67.60  \\
    3: CLIP-80 &	68.32 \\

    4: DCLIP (ICLR'23) & 68.03 \\
    5: Waffle CLIP (ICCV'23) & 68.34 \\
    6: CuPL (ICCV'23) & 69.62\\
    \midrule
    \rowcolor{tabhighlight}
    7: ProText-Attribute  & {{68.05}}\\
        \rowcolor{tabhighlight}
    8: ProText-80  &	{{68.48}}\\
        \rowcolor{tabhighlight}
    9: ProText-CuPL  &	\textbf{{70.22}}\\
    \bottomrule
    \end{tabular}
    }}
    \caption{With the same amount of text data, learning contextual prompts with text-only supervision improves CLIP performance in comparison to the prompt ensembling techniques. 
    }
    \label{table:different_text_data_motivation}
    \vspace{-3mm}
\end{SCtable}

%% file: tables/main_experiments/base_to_novel.tex
\begin{table}[!t]
    \small \centering
 \setlength{\tabcolsep}{4pt}
    \scalebox{0.75}[0.75]{
\begin{tabular}{l| ccc | ccc | ccc}
\toprule
{Dataset} & 
\multicolumn{3}{c}{CLIP \cite{radford2021learning}} & 
\multicolumn{3}{c}{CuPL \cite{zhou2022learning}} & 

 \multicolumn{3}{c}{ProText (Ours)} 

  \\
 & 
Base & 
Novel &
HM & 
Base & 
Novel &
HM & 
Base & 
Novel &
HM 

\\  \midrule

{ImageNet}       &72.43 &	68.14 	&70.22 &  74.30&	68.14 &71.09 &	\textbf{75.00}&\textbf{71.38 }&	\textbf{73.14 }\\

{Caltech101}     & 96.84 &94.00&95.40	&97.22 &	94.00 &95.58&	\textbf{98.06 }&\textbf{95.63}&	\textbf{96.83}\\

{OxfordPets}    & 91.17&97.26	&94.12&94.42&	97.26	&{95.82}&	\textbf{94.95}&\textbf{98.00}&	\textbf{96.45} \\

{StanfordCars}  &63.37 	&74.89&68.65&63.54&	74.89&68.75&	\textbf{64.54 }&\textbf{76.08}&	\textbf{69.84} \\

{Flowers102}    &72.08&77.80&74.83	&\textbf{74.36}&	77.80 &76.04&	\textbf{74.36}&\textbf{78.44}&	\textbf{76.35}\\

{Food101}      & 90.10&91.22&90.66&89.93&	91.22&90.57 &	\textbf{90.20}&\textbf{91.98}&	\textbf{91.08} \\
{Aircraft} &27.19&\textbf{36.29	}&31.09&30.61&	\textbf{36.29 }&\textbf{33.21} &	\textbf{30.91}&34.13&	32.44\\

{SUN397}       &69.36&75.35&72.23&76.02&	75.35&75.68&	\textbf{76.14}&\textbf{79.14}&	\textbf{77.61}\\

{DTD}           &53.24&59.90&56.37&62.85&	59.90 	&61.34&	\textbf{63.08}&\textbf{61.59}&	\textbf{62.33 }\\

{EuroSAT}       &56.48	&64.05&60.03&{59.64}&	64.05 &61.77&	\textbf{59.71}&\textbf{80.97}&	\textbf{68.73}\\

{UCF101}      &70.53&77.50&73.85&75.28&	77.50 &76.37&	\textbf{75.54}&\textbf{79.50}&	\textbf{77.47}\\

\midrule
 \rowcolor{tabhighlight} \textbf{{Average} }     &69.34 	&74.22 	&71.70&72.56&	74.22   &73.38&	\textbf{72.95}&\textbf{76.98 }&	\textbf{74.91} \\
\bottomrule
\end{tabular}
}
    \caption{\small\textnormal{\textbf{Base-to-novel setting.}} ProText enables the transferability of learned prompts to new classes and improves over CuPL \cite{pratt2023does}.}
    \label{tab:base-to-new}
\vspace{-4mm}
\end{table}

%% file: tables/main_experiments/cross_datasets.tex
\begin{SCtable*}[][!t]
    \tabstyle{4pt}
    \scalebox{0.868}{
    \begin{tabular}{l c ccccccccccc}
    \toprule
    & \textbf{Source} & \multicolumn{11}{c}{\textbf{Target}} \\ \cmidrule(lr){2-2} \cmidrule(lr){3-13}
    & \rotbox{ImageNet} & \rotbox{Caltech101} & \rotbox{OxfordPets} & \rotbox{StanfordCars} & \rotbox{Flowers102} & \rotbox{Food101} & \rotbox{Aircraft} & \rotbox{SUN397} & \rotbox{DTD} & \rotbox{EuroSAT} & \rotbox{UCF101} & \rotbox{\emph{Average}} \\
        \midrule
         \rowcolor{tabhighlight}\multicolumn{13}{c}{Methods utilizing labeled visual samples} \\
    CoOp & \textbf{{71.51}} & 93.70 & 89.14 & 64.51 & 68.71 & 85.30 & 18.47 & 64.15 & 41.92 & {{46.39}} & 66.55 & 63.88 \\
    Co-CoOp & 71.02 &{ 94.43} & {90.14} & 65.32 & {71.88} & 86.06 & 22.94 & \textbf{{67.36}} & 45.73 & 45.37 & 68.21 & 65.74 \\
    MaPLe & 70.72 & 93.53 & {{90.49}} & {65.57} &{ {72.23}} & {{86.20}} & {{{24.74}}} & 67.01 & {46.49} & {{{48.06}}} & {68.69} & {{66.30}}  \\
 PromptSRC & 71.27 & {93.60} & {{90.25}} & {65.70} & {70.25} & {{86.15}} & {{23.90}} & 67.10 & {46.87} & {{45.50}} & {68.75} & {{65.81}} \\
    \midrule
     \rowcolor{tabhighlight}\multicolumn{13}{c}{Zero-shot \& Prompt ensembling methods} \\
    CLIP & 66.72 & 92.98 & 89.13 & 65.29 & 71.30 & 86.11 & \textbf{24.90} & 62.59 & 44.56 & 47.84 & 66.83 & 65.15 \\
        CuPL & 69.62 & 92.98 & 89.13 & 65.29 & 71.30 & 86.11 & \textbf{24.90} & 62.59 & 44.56 & 47.84 & 66.83 & 65.15 \\
    \midrule
         \rowcolor{tabhighlight}\multicolumn{13}{c}{Prompt learning with text-only supervision} \\
 ProText (Ours) & 69.80 & \textbf{94.81} & \textbf{91.01} & \textbf{66.00} & \textbf{72.35} & \textbf{86.66} & {24.72} & 67.34 & \textbf{47.93} & \textbf{51.86} & \textbf{69.60} & \textbf{67.23} \\
    \bottomrule
    \end{tabular}}
        \caption{ \textbf{Cross-dataset transfer setting}. CuPL and CLIP perform same for cross-datasets as CuPL source data cannot transfer to cross-datasets. Image-based models are trained on 16-shot ImageNet samples. ProText employ same ImageNet data as CuPL for prompt learning. 
    }
    \label{tab:xd}
\end{SCtable*}

%% file: tables/main_experiments/domain_generalization.tex
\begin{table}[!t]

    \small \centering
 \setlength{\tabcolsep}{8pt}
    \scalebox{0.76}[0.76]{
    \begin{tabular}{l cccccc}
    \toprule
    & \textbf{Source} & \multicolumn{5}{c}{\textbf{Target}} \\ \cmidrule(lr){2-2} \cmidrule(lr){3-7}
     & ImageNet & -V2 & -S & -A & -R  & Avg.\\
    \midrule
                 \rowcolor{tabhighlight}\multicolumn{7}{c}{Methods utilizing labeled visual samples} \\
    CoOp &  \textbf{71.51} & \textbf{{64.20}} & 47.99  & 49.71  & 75.21  & {59.28} \\
    CoCoOp & 71.02 & {64.07} & 48.75 & 50.63 & 76.18 & {59.91}  \\
        MaPLe & 70.72  & {64.07} & 49.15  & 50.90 & 76.98 & {60.27}  \\
             \midrule
    \rowcolor{tabhighlight}\multicolumn{7}{c}{Zero-shot \& Prompt ensembling methods} \\
    CLIP &  66.72 & 60.83 & {46.15} & 47.77 & {73.96} & {57.18} \\
    CuPL &  69.62 & 63.27 & {49.02} & 50.72 & {77.05} & {60.01} \\

                     \midrule
                              \rowcolor{tabhighlight}\multicolumn{7}{c}{Prompt learning with text-only supervision} \\
 ProText (Ours) & 70.22 & {{63.54}} & \textbf{{49.45}} & \textbf{51.47}  & \textbf{{77.35}} & \textbf{{60.45}} \\
    \bottomrule
    \end{tabular}}\vspace{-0.9em}
        \caption{\textnormal{\textbf{Domain generalization.} }Prompt learning methods are trained on imageNet and evaluated on datasets with domain shifts.} 
    \label{tab:robustness}
\end{table}

%% file: tables/main_experiments/supervised_training_complete.tex
\begin{SCtable}[][!t]
    \small \centering
 \setlength{\tabcolsep}{6pt}
    \scalebox{0.83}[0.83]{
\begin{tabular}{l cccc }
\toprule
{Dataset}  & 
{CLIP}  & 
{CuPL}  &
{ProText} &
{$\Delta$}
\\  \midrule

{ImageNet}      & 66.72     &69.60	&\textbf{70.22} & \textcolor{MidnightBlue}{{+0.62}}\\

{Caltech101}    & 92.98     &94.32&\textbf{95.29} & \textcolor{MidnightBlue}{{+0.97}} \\

{DTD}           & 44.56       & {53.96} & \textbf{{54.02}} & \textcolor{MidnightBlue}{{+0.06}}\\

{EuroSAT}       & 47.84       & \textbf{60.27}&58.53& \textcolor{Bittersweet}{{-1.74}}\\

{StanfordCars}  & 65.29      & 65.95&\textbf{66.77} & \textcolor{MidnightBlue}{{+0.82}}\\

{Flowers102}    & 71.30      & 73.85&\textbf{74.42}& \textcolor{MidnightBlue}{{+0.57}}\\

{Aircraft}  & 24.90       & 27.66&\textbf{29.01}& \textcolor{MidnightBlue}{{+1.35}}\\

{SUN397}        & 62.59          & 69.00&\textbf{69.76} & \textcolor{MidnightBlue}{{+0.76}}\\

{OxfordPets}    & 89.13        &91.11&\textbf{92.72} & \textcolor{MidnightBlue}{{+1.61}}\\

{UCF101}        & 66.83        & 70.63&\textbf{71.45}& \textcolor{MidnightBlue}{{+0.82}}\\

{Food101}       & 86.11       &86.11&\textbf{86.68}& \textcolor{MidnightBlue}{{+0.57}} \\

\midrule
 \rowcolor{tabhighlight} \textbf{{Average} }      & 65.15       & 69.31&\textbf{69.90} &\textcolor{MidnightBlue}{+0.59} \\
\bottomrule
\end{tabular}
}\vspace{-6em}
    \caption{\small ProText results with text supervision on each dataset. We compare ProText with CLIP and CuPL. Gains of ProText over CuPL are shown in \textcolor{MidnightBlue}{blue}.}
    \label{tab:few_shot_experiments}
\vspace{-3em}
\end{SCtable}

%% file: tables/ablations/loss_ablation.tex
\begin{SCtable}[][!t]
    \small \centering
 \setlength{\tabcolsep}{0.1pt}
    \scalebox{0.85}[0.85]{
    \begin{tabular}{lc}
    \toprule
    Method  & ImageNet Top1. \\
    \midrule
    1: ProText-contrastive loss & {68.12} \\
    2: ProText- $L1$ loss & {69.96}	 \\
   \rowcolor{tabhighlight} 3: ProText-MSE loss  & \textbf{70.22}\\
    \bottomrule
    \end{tabular}}

    \caption{Ablation of choice of loss for contextual mapping. MSE loss provides highest results.}
    \label{tab:matching_loss_ablations}

\end{SCtable}

%% file: tables/ablations/LLM_ablation.tex
\begin{SCtable}[][!t]
    \small \centering
 \setlength{\tabcolsep}{0.0pt}
    \scalebox{0.85}[0.85]{
    \begin{tabular}{lc}
    \toprule
    Method  & ImageNet Top1 \\
    \midrule
    1: ProText-80 templates & 68.48	 \\
    2: ProText-Alpaca & 67.10 \\
   \rowcolor{tabhighlight} 3: ProText-GPT-3  & \textbf{70.22}\\
    \bottomrule
    \end{tabular}}
    \caption{Effect on performance with different text data for training. GPT-3 text data show highest results.}
    \label{tab:LLM_ablation}
\end{SCtable}

%% file: tables/ablations/adaptor_ablation.tex
\begin{SCtable}[][!t]
    \small \centering
 \setlength{\tabcolsep}{5pt}
    \scalebox{0.80}[0.80]{
    \begin{tabular}{lc}
    \toprule
    Method  & ImageNet Top1. \\
    \midrule
    1: Linear Adaptor & 69.36 \\
    2: MLP Adaptor & 69.24	\\
   \rowcolor{tabhighlight} 3: Prompt Learning  & \textbf{70.22}\\
    \bottomrule

    \end{tabular}}
    \caption{Ablation on the choice of mapping network. Prompt Learning shows optimal performance.}
    \label{tab:adaptor_ablation}
\end{SCtable}

%% file: tables/main_experiments/prediction_score_analysis.tex
\begin{table}[!t]
\setlength{\tabcolsep}{3pt}
\centering
\resizebox{1\linewidth}{!}{%
\begin{tabular}{lcccc|cccc}
\toprule
 & \multicolumn{4}{c}{Correct class confidence (\%)  {$\uparrow$}} & \multicolumn{4}{c}{Incorrect class confidence (\%) {$\downarrow$}}  \\
 Method & DTD & SUN & Caltech & UFC & DTD & SUN & Caltech & UFC \\
\midrule
CLIP & 30.5 & 49.3 & 84.5  & 56.4 &  1.51& {0.13} & {0.16}& {0.44}\\
 \rowcolor{tabhighlight} ProText & \textbf{33.1	}& \textbf{54.2} & \textbf{89.1}  & \textbf{59.5} &  \textbf{1.45}& \textbf{0.12} & \textbf{0.11} & \textbf{0.40}\\
\bottomrule
\end{tabular}}
\caption{ Confidence score analysis: ProText trained on ImageNet improves its logit confidence for correct classes in unseen datasets.}
\label{tab:confidence_analysis}
\end{table}

%% file: sec/3_finalcopy.tex
\vspace{-0.5em}
\section{Conclusion}
Prompt learning and LLM-based ensembling are effective techniques to improve CLIP's generalization. However, prompt learning often requires labeled images, which is less practical, while LLM-based ensembling methods are dominantly class-specific and not directly transferable to new classes. To address these challenges, we propose a new direction to adapt CLIP by learning generalized prompts with text-only supervision, without relying on visual data. We introduce a training strategy for prompts to learn a mapping function that embeds rich contextual knowledge from LLM text data within the prompts. The context learned by these prompts transfers well to unseen classes and datasets, potentially reducing the LLM prompt engineering and serving cost. We perform extensive evaluations on four benchmarks where our text-only approach performs favorably well over previous methods, including those utilizing labeled images.\vspace{1em}

\noindent \textbf{Acknowledgements:} We would like to thank Hanan Ghani and Jameel Hassan for their help in downloading datasets. We also thank Muhammad Jehanzeb Mirza for providing Alpaca LLM prompt data for ablation experiments.

%% file: sec/X_suppl.tex
\clearpage
\setcounter{page}{1}
\appendix
\maketitlesupplementary

The following sections provide supplementary material for our main paper. This includes additional analysis and comparison experiments, implementation details, and specifics of our text-to-text data used for training.  The contents are organized as follows:
\begin{itemize}

        \item Additional analysis and comparison experiments (Sec. \ref{appendix:additional_experiments})
         \item Additional Implementation details (Sec. \ref{appendix:additional_implementation_details})
    \item Details on Text-Only Data (Sec. \ref{appendix:text_data_details})

\end{itemize}

\section{Additional Experiments}
\label{appendix:additional_experiments}

\subsection{Additional Analysis.} 
Here we provide additional analysis experiments for our ProText technique. 

\begin{figure}[!h]
    \centering
    \includegraphics[width=0.95\columnwidth]{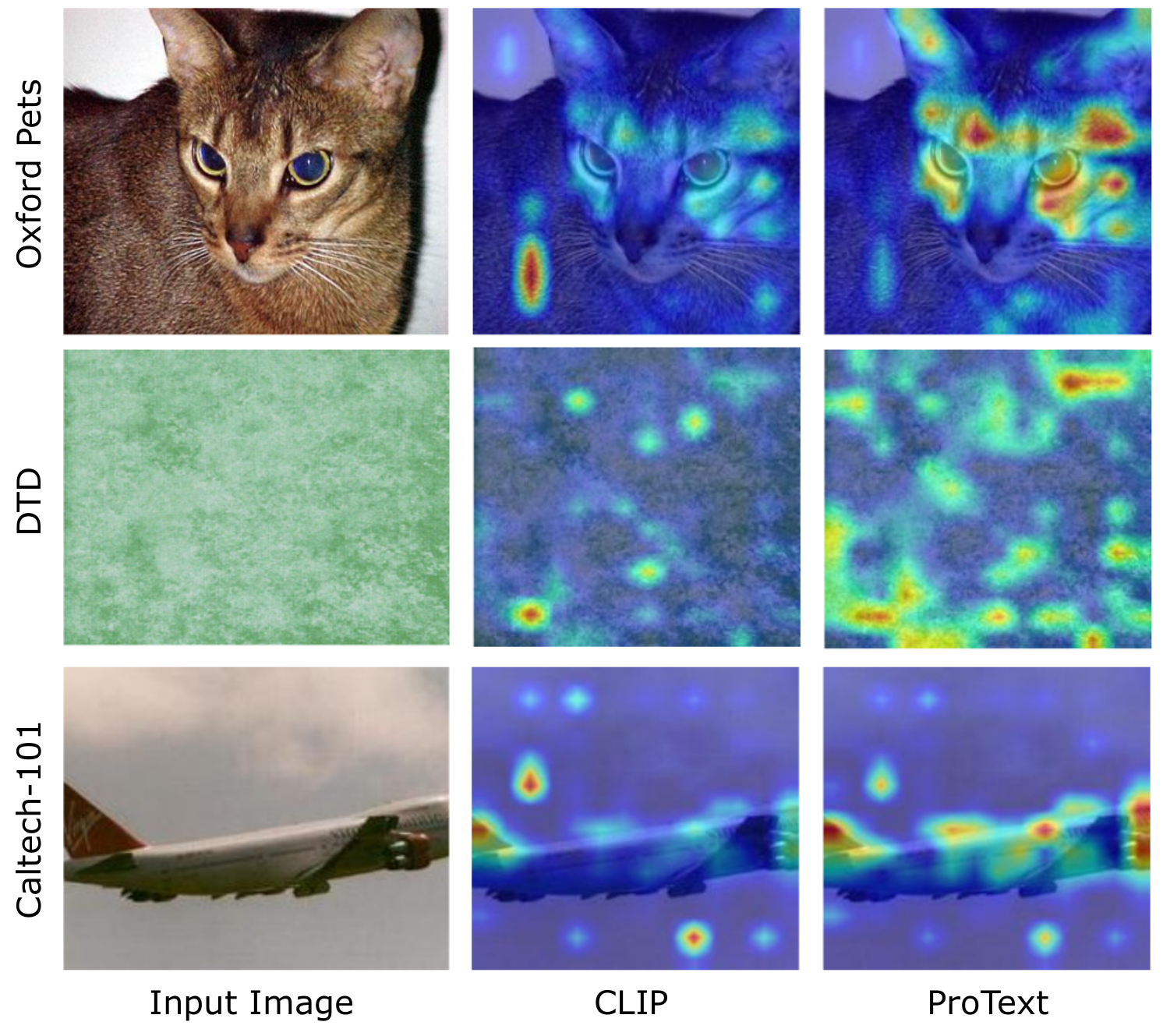}
\caption{Attention map visualizations for CLIP and ProText for cross-datasets. ProText is trained on ImageNet-1k text-only data.}
  \label{fig:attn_map_visualization}
\end{figure}

\noindent \textbf{Qualitative Analysis.}
In order to understand the transferability of ProText prompts across new datasets, we visualize attention maps in Fig. \ref{fig:attn_map_visualization}. Specifically, we employ ProText prompts learned on ImageNet-1k text-only dataset and transfer it to cross-datasets. We observe that ProText tends to focus to relevant image features while reducing its attention towards spurious features as shown in Oxford Pets and Caltech-101 images. In case of texture image from DTD, ProText shows more global attention on the texture portion of the image which is crucial in recognizing the correct texture due to the fine-grained nature of texture classes. This suggest that ProText can learn complementary contextual features, which steers CLIP for better transferability towards new datasets without relying on visual samples.

\noindent \textbf{Towards interpreting ProText prompts.}
Our main experiments in \autoref{sec:ablation_studies} demonstrated that ProText trained on ImageNet-1k text dataset performs favorably well across cross-datasets. Here we are interested in studying how the ProText prompt vectors are interpreted in natural language. Specifically, we searched for words in CLIP vocabulary that are closest to the learned prompts using Euclidean distance in the embedding space. The results in Table \ref{tab_appendix:interpret_prompts} show the nearest (valid) word for ProText prompts across different transformer layers. Note that these words may note concretely correspond to the learned prompts as we could only select nearest ones. We observe that the represented words are diverse containing connecting words that are common in web captions such as "a," "for," and "onto". Additionally, since CLIP uses a BPE representation for tokenization, several subwords appear among the nearest words, such as "sin," "ced," and "banan." These subwords can collectively contribute to strong context priors, such as deriving "banana" from "banan," "Mercedes" from "ced," and "casino" from "sin," which may be potentially relevant for downstream datasets like SUN397 and Stanford Cars. At the same time, some words do not appear to contribute much for context enhancement such as "ilwx", "curfew" etc. In summary, similar to the findings in \cite{zhou2022learning}, the learned vectors may encompass word representations not explicitly present in the existing vocabulary.

\input{tables/main_experiments/interpret_prompt}

\subsection{Additional comparisons with WaffleCLIP.} We present additional comparisons between ProText and WaffleCLIP \cite{roth2023waffling} approach. WaffleCLIP employs prompt ensembling by introducing random descriptors and characters alongside class names. Specifically, we perform a comparison with a WaffleCLIP-Concept variant, which incorporates high-level dataset concepts in its text prompts, such as $\texttt{`a photo of a flower: a CLS } \texttt{'}$ for OxfordFlowers \cite{nilsback2008automated}. Further details on the WaffleCLIP framework and its variants can be found in \cite{roth2023waffling}.

\input{tables/supplementary/cross_dataset_waffle_clip}

\noindent \textbf{Cross-dataset transfer.} For cross-dataset transfer settings, all methods only utilize ImageNet source dataset LLM prompt information. The results are shown in Tab. \ref{tab:xd_supplementary}. CuPL shows the same performance as CLIP for cross-datasets as class-specific descriptions for new datasets are not available in this setting. Overall, WaffleCLIP uses random descriptors which leads to improvements over CLIP and CuPL. In contrast, ProText with text-only training with ImageNet-1k LLM templates shows consistent improvements over WaffleCLIP by surpassing on 9/10 cross-datasets and leads to the averaged accuracy of 67.23\% in the challenging cross-dataset transfer setting.

\noindent \textbf{Text-only supervised setting.} We additionally compare WaffleCLIP in text-only supervised setting. As shown in Tab. \ref{tab_appendix:text_only_waffle_clip}, WaffleCLIP improves over CLIP but lags behind CuPL as it only relies on high-level dataset concepts and random descriptors. CuPL uses class-specific LLM descriptions for prompt ensembling and shows improved results. In contrast to these approaches, ProText adopts a learning-based approach using text data and shows the highest performance by surpassing both WaffleCLIP and CuPL in 10/11 datasets. This suggests that text-only prompt learning can serve as a better alternative to training-free prompt ensembling methods.

\input{tables/supplementary/text_only_supervised_waffle_clip}

\subsection{Comparison with image-supervised methods.}
We show additional comparisons of ProText with image-supervised methods in terms of generalization performance.  In base to novel class generalization setting, we include prompt learning methods utilizing 16-shot image data where we mainly focus on novel class performance for comparison. In text-only supervised setting, we compare ProText with few-shot image supervised methods including CLIP Linear Probe, CoOp, and CoCoOp, which are trained up to 2-shot data.

\noindent \textbf{Unseen class generalization.} All methods are trained on seen classes of each dataset and we specifically analyze their performance on unseen classes to study generalization. Results are shown in Tab. \ref{tab_appendix:b2b_only_unseen_waffle_clip}. Image-supervised prompt learning methods utilize 16-shot base-class labeled data and demonstrate improved accuracy for novel classes.  For example, the previous state-of-the-art method, PrompSRC, achieves a substantial accuracy of 70.73\% on ImageNet for novel classes. In comparison, ProText, leveraging text-only data, shows an improvement of +0.65\% against PromptSRC for novel classes on ImageNet. In summary, ProText consistently outperforms PromptSRC on 9 out of 11 datasets for novel classes, leading to the highest novel class accuracy of 76.98\% averaged over 11 datasets.

\input{tables/supplementary/b2n_image_comparision}

\noindent \textbf{Supervised setting.}
In Tab. \ref{tab_appendix:few_shot_experiments}, compare ProText with few-shot image-supervised methods including CLIP Linear Probe, CoOp, and CoCoOp. ProText shows improved averaged performance over 1 \& 2 shot Linear Probe. Similarly, ProText without using any images for training improves on most datasets against CoOp and CoCoOp trained with 1 and 2 shots. ProText, without using any images for training, outperforms CoOp and CoCoOp trained with 1 and 2 shots on most datasets. This suggests that text-only training can be considered an effective alternative approach to image-supervised methods under extreme low-data regimes.

\input{tables/supplementary/text_only_with_few_shot_imagemethods}

\subsection{Additional ablation studies.}
We present additional ablation experiments conducted on ProText as outlined below.

\noindent \textbf{Combining prompt ensembling and prompt learning.}
In our ProText approach, learnable prompts for inference are trained on text data. Here, we explore an alternative experiment by averaging the text features with ProText-learned prompts and text features of LLM templates obtained via prompt ensembling. Specifically, we average the LLM prompt features (e.g., CuPL features) and ProText features for the same classes to study if prompt learning and prompt ensembling could be complementary. The results are shown in Table \ref{tab:appendix_averaging_protext_and_cuple}. Combining ProText and CuPL features leads to marginal improvement compared to ProText alone. We conjecture that since ProText uses the same LLM template data to learn prompts, the LLM template features and ProText features might not be strongly complementary.

\input{tables/supplementary/ablation_protext_with_raw_ensembling}

\section{Additional Implementation details}
\label{appendix:additional_implementation_details}
\noindent \textbf{{Training details.}} For training ProText, we use a publically available CLIP ViT-B/16 model from OpenAI \cite{radford2021learning}. Language prompts for each training are initialized with `a photo of a" for the first layer and randomly initialized for the remaining transformer layers of the text encoder of CLIP. All models are trained using the AdamW optimizer on a single 16-GB V100 GPU. For cross-dataset and domain generalization benchmarks, we train ProText using $T=4$ and $T=16$ language prompts, respectively, for 10 and 200 epochs, respectively. The warm-up epochs are set to 5 during training.

 \noindent As text data from LLMs varies in quality and size across datasets, we have observed that training ProText on each dataset requires custom training configurations to achieve the best performance. Therefore, ProText employs optimal prompt length and epoch configuration for each dataset. The optimal training configurations are obtained through the validation splits of each dataset.

 \noindent \textbf{Base-to-novel generalization setting.}
In Tab. \ref{tab:b2n_hyperparameters}, we show the hyperparameters used for training models in base-to-novel generalization settings. We use a learning rate of 0.03 for all datasets except UCF101, FOOD101, and Oxford-Flowers where learning rate of 0.0025 is used.

\input{tables/supplementary/base_to_novel_hyper_parameters}

 \noindent \textbf{Text-only supervised setting.} For our comparison with CuPL \cite{pratt2023does} in Table \ref{tab_appendix:few_shot_experiments}, ProText models are trained using the same LLM text data as utilized by CuPL. Hyperparameter values are shown in Table \ref{tab:fs_hyperparameters}. All models are trained using a learning rate of 0.03, except for UCF101, EuroSAT, and Oxford-Flowers, where a learning rate of 0.0025 is used.

 \input{tables/supplementary/fully_supervised_hyper_parameters}

\section{Details on Text-Only Data}
As discussed in \autoref{text_data_curation_para}, our  ProText approach relies on text-only data ($\mathcal{D}{ \textsc{prompt}}$) curated from Language Models (LLMs) for training its language prompts. Here, we provide additional details on the curation of text-only data. Specifically, we first provide information on the text queries used as input to LLMs for generating prompts, followed by qualitative examples of $\mathcal{D}_{ \textsc{prompt}}$.
\label{appendix:text_data_details}
\subsection{Queries to LLMs to curate Text-Only Data}
Following \cite{pratt2023does}, we obtain class descriptions from LLMs by providing various queries as inputs. Specifically, we utilize queries termed as \textit{Full prompts} by CuPL \cite{pratt2023does}. For instance, to generate class descriptions of ImageNet-1k classes, we prompt GPT-3 with the following 5 queries:

\begin{itemize}
        \item $\texttt{`Describe what a(n) CLS looks like.} \texttt{'}$
    \item $\texttt{`How can you identify a(n) CLS? } \texttt{'}$
    \item $\texttt{`What does a(n) look like? } \texttt{'}$
        \item $\texttt{`Describe an image from the internet of a(n) CLS.}\texttt{'}$
            \item $\texttt{`A caption of an image of a(n) CLS.} \texttt{'}$
\end{itemize}

Here, $\texttt{CLS}$ denotes the class names present in the dataset. After generating LLM class descriptions, we associate all descriptions of the same class with its class-name template given as $\texttt{`A photo of a CLS'}$. This results in our text-only training data $\mathcal{D}_{ \textsc{prompt}}$ with text-to-text mapping pairs used to train ProText. Refer to \cite{pratt2023does} for LLM queries of other datasets used to generate class-specific descriptions. For standardized comparisons, we use publicly available CuPL data and generate descriptions for datasets not provided by CuPL.

\subsection{Qualitative examples}
As LLMs are pre-trained on internet-scale text corpora, they possess the capability of generating diverse and high-quality descriptions and captions for different class categories, resulting in high-quality text outputs. Below we show some examples of $\mathcal{D}_{ \textsc{prompt}}$ text-to-text pairs for the ImageNet-1k dataset.

\noindent \textbf{Class: Tench }

\noindent Class-name template: $\texttt{`A photo of a Tench'}$

\noindent Associated LLM descriptions:
\begin{itemize}
        \item \texttt{`A tench is a freshwater fish with a dark green back and light-colored sides.'}
        \item \texttt{`A tench looks like a freshwater fish with a dark olive-green back, fading to yellowish-brown on the sides.'}

                \item \texttt{`Tench are a freshwater fish that can grow up to 70cm long! They have olive-brown skin with dark spots, and their meat is white and firm.'}

                        \item \texttt{`This image shows a large, dark green tench swimming in a pond.'}
\end{itemize}

\noindent \textbf{Class: bath towel }

\noindent Class-name template: $\texttt{`A photo of a bath towel'}$

\noindent Associated LLM descriptions:
\begin{itemize}
        \item \texttt{`A bath towel typically has a loops on one side and a smooth surface on the other.'}
        \item \texttt{`A bath towel is a rectangular piece of fabric, usually Cotton, that is used to dry oneself after a bath or shower.'}

                \item \texttt{`The image is of a white bath towel with a blue and green stripes.'}

                        \item \texttt{`A fluffy white bath towel draped over a towel rack.'}
\end{itemize}

\noindent \textbf{Class: sandal }

\noindent Class-name template: $\texttt{`A photo of a sandal'}$

\noindent Associated LLM descriptions:
\begin{itemize}
        \item \texttt{`A sandal is a shoe typically made of leather or synthetic material that has an open toe and a strap or straps that go around the foot or up the ankle.'}
        \item \texttt{`A sandal is usually a flat shoe with a strap that goes around the foot or ankle.'}

                \item \texttt{`This sandal is from the ancient Egyptian city of Thebes.'}

                        \item \texttt{`When you are looking to identify a sandal, the first place to start is by looking at the features of the shoe.'}
\end{itemize}

%% file: tables/main_experiments/interpret_prompt.tex
\begin{table}[!t]
    \small \centering
 \setlength{\tabcolsep}{4pt}
    \scalebox{0.95}[0.95]{
\begin{tabular}{l cccc }
\toprule
{Layer \#}  & 
{CTX 1}  & 
{CTX 2}  &
{CTX 3}  &
{CTX 4} 
\\  \midrule

{1}      & a    &a& for& onto\\

{2}    & bi     &paper&erup&believes  \\

{3}           & ilwx       & ered & emon & enclosure \\

{4}       & devoted       & fly&ced&hair\\

{5}  & sin      &tous&cona&emor \\

{6}    & foto      & unwanted&swagg&curfew\\

{7}  & banan       & lift&knob&maz\\

{8}        & slow         &commuter&helene&nuff \\

{9}    & chevron        &rear&crepe&opi \\

\bottomrule
\end{tabular}
}
    \caption{\small Illustration of nearest words in CLIP word vocabulary against ProText prompts in different transformer layers. ProText prompts are trained on ImageNet-1k LLM prompt data.}
    \label{tab_appendix:interpret_prompts}
\end{table}

%% file: tables/supplementary/cross_dataset_waffle_clip.tex
\begin{table*}[!ht]
    \tabstyle{4pt}
    \scalebox{0.868}{
    \begin{tabular}{l c ccccccccccc}
    \toprule
    & \textbf{Source} & \multicolumn{11}{c}{\textbf{Target}} \\ \cmidrule(lr){2-2} \cmidrule(lr){3-13}
    & \rotbox{ImageNet} & \rotbox{Caltech101} & \rotbox{OxfordPets} & \rotbox{StanfordCars} & \rotbox{Flowers102} & \rotbox{Food101} & \rotbox{Aircraft} & \rotbox{SUN397} & \rotbox{DTD} & \rotbox{EuroSAT} & \rotbox{UCF101} & \rotbox{\emph{Average}} \\
    \midrule
     \rowcolor{tabhighlight}\multicolumn{13}{c}{Zero-shot \& Prompt ensembling methods} \\

    CLIP & 66.72 & 92.98 & 89.13 & 65.29 & 71.30 & 86.11 & \textbf{24.90} & 62.59 & 44.56 & 47.84 & 66.83 & 65.15 \\
        CuPL & 69.62 & 92.98 & 89.13 & 65.29 & 71.30 & 86.11 & \textbf{24.90} & 62.59 & 44.56 & 47.84 & 66.83 & 65.15 \\
                WaffleCLIP-Concept & 68.34 & 94.01 & 89.57 & 63.42 & 72.00 & \textbf{86.84} & 24.49 & 66.17 & 45.15 & 47.74 & 67.96 & 65.74 \\

    \midrule
         \rowcolor{tabhighlight}\multicolumn{13}{c}{Prompt learning with text-only supervision} \\
 ProText (Ours) & \textbf{69.80} & \textbf{94.81} & \textbf{91.01} & \textbf{66.00} & \textbf{72.35} & {86.66} & {24.72} & \textbf{67.34} & \textbf{47.93} & \textbf{51.86} & \textbf{69.60} & \textbf{67.23} \\

    \bottomrule
    \end{tabular}}
        \caption{ \textbf{Cross-dataset transfer setting}. Results comparison of ProText with CLIP, CuPL, and Waffle-CLIP. ProText overall shows consistent improvements over LLM-based prompt ensembling methods.
    }
    \label{tab:xd_supplementary}
\end{table*}

%% file: tables/supplementary/text_only_supervised_waffle_clip.tex
\begin{table}[!h]
    \small \centering
 \setlength{\tabcolsep}{4pt}
    \scalebox{0.95}[0.95]{
\begin{tabular}{l ccccc }
\toprule
{Dataset}  & 
{CLIP}  & 
{CuPL}  &
{WaffleCLIP-C}  &
{ProText} &
{$\Delta$}
\\  \midrule

{ImageNet}      & 66.72     &69.60	&68.34&\textbf{70.22} & \textcolor{MidnightBlue}{{+0.62}}\\

{Caltech101}    & 92.98     &94.32&94.01&\textbf{95.29} & \textcolor{MidnightBlue}{{+0.97}} \\

{DTD}           & 44.56       & {53.96} &45.15 & \textbf{{54.04}} & \textcolor{MidnightBlue}{{+0.06}}\\

{EuroSAT}       & 47.84       & \textbf{60.27}&47.74&58.53& \textcolor{Bittersweet}{{-1.74}}\\

{StanfordCars}  & 65.29      & 65.95&63.42&\textbf{66.77} & \textcolor{MidnightBlue}{{+0.82}}\\

{Flowers102}    & 71.30      & 73.85&72.00&\textbf{74.42}& \textcolor{MidnightBlue}{{+0.57}}\\

{Aircraft}  & 24.90       & 27.66&24.49&\textbf{29.01}& \textcolor{MidnightBlue}{{+1.35}}\\

{SUN397}        & 62.59          & 69.00&66.17&\textbf{69.76} & \textcolor{MidnightBlue}{{+0.76}}\\

{OxfordPets}    & 89.13        &91.11&89.57&\textbf{92.72} & \textcolor{MidnightBlue}{{+1.61}}\\

{UCF101}        & 66.83        & 70.63&67.96&\textbf{71.45}& \textcolor{MidnightBlue}{{+0.82}}\\

{Food101}       & 86.11       &86.11&\textbf{86.84}&{86.68}& \textcolor{MidnightBlue}{{+0.57}} \\

\midrule
 \rowcolor{tabhighlight} \textbf{{Average} }      & 65.15       & 69.31&65.97&\textbf{69.90} &\textcolor{MidnightBlue}{+0.59} \\
\bottomrule
\end{tabular}
}
    \caption{\small ProText results with text supervision on each dataset. We compare ProText with CLIP and CuPL and WaffleCLIP-Concept. Gains of ProText over CuPL are shown in \textcolor{MidnightBlue}{blue}.}
    \label{tab_appendix:text_only_waffle_clip}
\end{table}

%% file: tables/supplementary/b2n_image_comparision.tex
\begin{table}[!h]
    \small \centering
\tablestyle{-12pt}{1.1}
\addtolength{\tabcolsep}{+14pt}
    \scalebox{0.95}[0.95]{
\begin{tabular}{l cc|ccccc }
\toprule
{Dataset}  & 
{CuPL}  & 
{ProText}  &
{CoOp} &
{CoCoOp} &
{MaPLe} &
{PromptSRC} &
{$\Delta$} 
 \\
 &  \cite{pratt2023does} & Ours & \cite{zhou2022learning}  & \cite{zhou2022conditional}  & \cite{khattak2023maple}  & \cite{khattak2023self} & 
\\  \midrule

{ImageNet}      & 68.14    &\textbf{71.38 }	&67.88&70.43 &70.54 & 70.73& \textcolor{MidnightBlue}{{+3.2}}\\

{Caltech101}    & 94.00    &\textbf{95.63}&89.81&93.81&94.36& 94.03& \textcolor{MidnightBlue}{{+1.6}} \\

{DTD}           & 59.90       & 61.59&41.18 & 56.00 & 59.18&\textbf{62.97}&\textcolor{MidnightBlue}{{+1.7}}\\

{EuroSAT}       &  64.05       & \textbf{80.97}&54.74&60.04&73.23&73.90& \textcolor{MidnightBlue}{{+17}}\\

{StanfordCars}  & 74.89      & \textbf{76.08}&60.40&73.59 &74.00 & 74.97& \textcolor{MidnightBlue}{{+1.2}}\\

{Flowers102}    & 77.80      & \textbf{78.44}&59.67 &71.75& 72.46&76.50&\textcolor{MidnightBlue}{{+0.6}}\\

{Aircraft}  & {36.29 }      & 34.13 &22.30&23.71& 35.61 &\textbf{37.87}&\textcolor{Bittersweet}{{-2.2}}\\

{SUN397}        & 75.35         & \textbf{79.14} &65.89&76.86&78.70 &78.47& \textcolor{MidnightBlue}{{+3.8}}\\

{OxfordPets}    & 97.26        &\textbf{98.00}&95.29&97.69&97.76& 97.30& \textcolor{MidnightBlue}{{+0.7}}\\

{UCF101}        & 77.50       & \textbf{79.50}&56.05&73.45&78.66  &78.80& \textcolor{MidnightBlue}{{+2.0}}\\

{Food101}       & 91.22       &91.98 &82.26 &{91.29 }& \textbf{92.05 }&91.53& \textcolor{MidnightBlue}{{+0.8}} \\

\midrule
 \rowcolor{tabhighlight} \textbf{{Average} }      & 74.22       & \textbf{76.98}&63.22 &71.69 &75.14 &76.10&\textcolor{MidnightBlue}{+2.8} \\
\bottomrule
\end{tabular}
}
    \caption{\small \textbf{Novel-class generalization comparison.}We compare ProText with prompt ensembling and image-supervised methods on unseen class performance in base-to-novel class generalization setting. Gains of ProText over CuPL are shown in \textcolor{MidnightBlue}{blue}.}
    \label{tab_appendix:b2b_only_unseen_waffle_clip}
\end{table}

%% file: tables/supplementary/text_only_with_few_shot_imagemethods.tex
\begin{table*}[!t]
    \small \centering
 \setlength{\tabcolsep}{7pt}
    \scalebox{1}[1]{
\begin{tabular}{l| ccc | cc cc cc}
\toprule
\multirow{2}{*}{Dataset}  & 
\multirow{2}{*}{CLIP}  & 
\multirow{2}{*}{CuPL}  &
\multirow{2}{*}{ProText}& 
 \multicolumn{2}{c}{Linear Probe} & 
 \multicolumn{2}{c}{CoOp} & 

 \multicolumn{2}{c}{CoCoOp} 

  \\
 & 
 & 
&
& 
{$K$=1} & 
{$K$=2} & 
{$K$=1} & 
{$K$=2} & 
{$K$=1} & 
{$K$=2} 

\\  \midrule

{ImageNet}      & 66.70     &69.62	&\textbf{70.22}&32.13&	44.88 &   66.33&	67.07 &69.43&	69.78 \\

{Caltech101}    & 92.98     &94.32&\textbf{95.29} & 79.88	&89.01&92.60&	93.07 &93.83&	94.82 \\

{DTD}           & 44.56       & {53.96}&\textbf{{54.04}}&34.59&40.76&50.23&	53.60 	&48.54&	52.17 \\

{EuroSAT}       & 47.84       & \textbf{60.27}&58.53&49.23	&61.98	&54.93&	65.17 &55.33&	46.74 \\

{StanfordCars}  & 65.29      & 65.95&\textbf{66.77}&35.66	&50.28&67.43&	70.50&67.22&	68.37 \\

{Flowers102}    & 71.30      & 73.85&\textbf{74.42}&69.74	&85.07	&77.53&	87.33 &72.08&	75.79 \\

{Aircraft}  & 24.90       & 27.66&\textbf{29.01}&19.61	&26.41&21.37&	26.20  &12.68  &	15.06\\

{SUN397}        & 62.59          & 69.00&\textbf{69.76}&41.58	&53.70&66.77&	66.53 &68.33&	69.03 \\

{OxfordPets}    & 89.13        &91.11&\textbf{92.72}& 44.06	&58.37&90.37&	89.80	&91.27&	92.64 \\

{UCF101}        & 66.83        & 70.63&\textbf{71.45}&53.66	&65.78&71.23&	73.43 &70.30&	73.51 \\

{Food101}       & 86.11       &86.11&\textbf{86.68}& 43.96	&61.51&84.33&	84.40&85.65&	86.22 \\

\midrule
 \rowcolor{tabhighlight} \textbf{{Average} }      & 65.15       & 69.31&\textbf{69.90}&45.83	&57.98	&67.56&	70.65    &66.79&	67.65 \\
\bottomrule
\end{tabular}
}
    \caption{\small ProText results with text supervision on each dataset. We compare ProText with CLIP \cite{radford2021learning}, CuPL \cite{pratt2023does}  and image supervised Linear Probe \cite{radford2021learning}, CoOp \cite{zhou2022learning} and CoCoOp \cite{zhou2022conditional} methods.}
    \label{tab_appendix:few_shot_experiments}
\vspace{-2mm}
\end{table*}

%% file: tables/supplementary/ablation_protext_with_raw_ensembling.tex
\begin{SCtable}[][!t]
    \small \centering
 \setlength{\tabcolsep}{5pt}
    \scalebox{0.80}[0.80]{
    \begin{tabular}{lc}
    \toprule
    Method  & ImageNet Top1. \\
    \midrule
    1: CuPL & 69.62 \\
    2: ProText & 70.22	\\
    3: Ensembling: ProText + CuPL  & \textbf{70.28}\\
    \bottomrule

    \end{tabular}}
    \caption{Ablation on combining CuPL and ProText text features.}
    \label{tab:appendix_averaging_protext_and_cuple}
\end{SCtable}

%% file: tables/supplementary/base_to_novel_hyper_parameters.tex
\begin{table}[!h]
    \tabstyle{3pt}
    \scalebox{0.9}{
    \begin{tabular}{l cccccccccccc}
    \toprule
H.parameter & \rotboxsub{ImageNet} & \rotboxsub{Caltech101} & \rotboxsub{OxfordPets} & \rotboxsub{StanfordCars} & \rotboxsub{Flowers102} & \rotboxsub{Food101} & \rotboxsub{Aircraft} & \rotboxsub{SUN397} & \rotboxsub{DTD} & \rotboxsub{EuroSAT} & \rotboxsub{UCF101} \\
        \midrule
  Epochs & 30 & 30 & 50 & 30 & 150 & 50 & 200 & 30 & 200 & 30 & 20 \\
    \# Prompts ($T$) & 4 & 8 & 4 & 8 & 4 & 8 & 4 & 8 & 4 & 16 & 16 \\
    \bottomrule
    \end{tabular}}

    \caption{Hyper-parameters setting used for base-to-novel generalization setting. Optimal configuration is set using validation splits of each dataset.}
    \label{tab:b2n_hyperparameters}
\end{table}

%% file: tables/supplementary/fully_supervised_hyper_parameters.tex
\begin{table}[!h]
    \tabstyle{3pt}
    \scalebox{0.9}{
    \begin{tabular}{l cccccccccccc}
    \toprule
H.parameter & \rotboxsub{ImageNet} & \rotboxsub{Caltech101} & \rotboxsub{OxfordPets} & \rotboxsub{StanfordCars} & \rotboxsub{Flowers102} & \rotboxsub{Food101} & \rotboxsub{Aircraft} & \rotboxsub{SUN397} & \rotboxsub{DTD} & \rotboxsub{EuroSAT} & \rotboxsub{UCF101} \\
        \midrule
  Epochs & 200 & 30 & 50 & 20 & 300 & 30 & 200 & 200 & 200 & 300 & 100 \\
    \# Prompts ($T$) & 16 & 16 & 4 & 8 & 4 & 16 & 4 & 16 & 16 & 4 & 8 \\
    \bottomrule
    \end{tabular}}

    \caption{Hyper-parameters used for text-only supervised setting.} \vspace{-0.5em}
    \label{tab:fs_hyperparameters}
\end{table}